%% file: main.tex
\documentclass[runningheads]{llncs}
\newcommand{\ourwork}{Bench2\allowbreak{Drive}-Speed}
\newcommand{\ourdataset}{Customized\allowbreak Speed\allowbreak Dataset}
\newcommand{\btd}{Bench2Drive1K}
\newcommand{\btdlong}{Bench2Drive1K\allowbreak-Long}
\newcommand{\btdshort}{Bench2Drive1K\allowbreak-Short}
\newcommand{\expertdata}{Expert2.1k}
\newcommand{\virtuallong}{Virtual2.1k\allowbreak-Long}
\newcommand{\virtualshort}{Virtual2.1k\allowbreak-Short}
\newcommand{\ouragent}{PDM\allowbreak-Lite\allowbreak-Speed}
\newcommand{\ourbaseline}{TCP\allowbreak-Speed}
\newcommand{\carla}{CARLA}
\newcommand{\parhead}[1]{\vspace{0.5em}\noindent\textbf{#1}\ }

\usepackage{xcolor}

\newcommand{\g}{\color{gray!80}}

\usepackage{tabularx}
\usepackage{makecell}
\usepackage{wrapfig}
\usepackage{multirow}
\usepackage{amssymb}
\usepackage[table]{xcolor}
\usepackage{pifont}
\usepackage{bbding}

\usepackage{eccv}

\usepackage{eccvabbrv}

\usepackage{graphicx}
\usepackage{booktabs}
\usepackage{caption}

\usepackage{enumitem}
\usepackage{mytitlesec}
\titlespacing\section{5pt}{5pt plus 5pt minus 5pt}{5pt plus 5pt minus 5pt}
\titlespacing\subsection{5pt}{5pt plus 5pt minus 5pt}{5pt plus 5pt minus 5pt}
\titlespacing\subsubsection{5pt}{5pt plus 5pt minus 5pt}{5pt plus 5pt minus 5pt}
\usepackage[accsupp]{axessibility}  

\usepackage{hyperref}

\usepackage{orcidlink}

\begin{document}

\title{Can Users Specify Driving Speed? \\ \ourwork: Benchmark and Baselines for Desired-Speed Conditioned Autonomous Driving} 

\titlerunning{\ourwork}

\author{Yuqian Shao\inst{1} \and
Xiaosong Jia\inst{2}\Envelope \and
Langechuan Liu\inst{3} \and
Junchi Yan\inst{1}\Envelope}

\authorrunning{Y.~Shao et al.}

\institute{Sch. of Computer Science \& Sch. of Artificial Intelligence, \\ Shanghai Jiao Tong University \and Institute of Trustworthy Embodied AI (TEAI), Fudan University \and NVIDIA \\
\Envelope \, Correspondence Authors\\
}

\maketitle

\begin{figure}[h]
    \centering
    \begin{minipage}{0.8\textwidth}
        \centering
        \vspace{-20pt}
        \includegraphics[width=\linewidth]{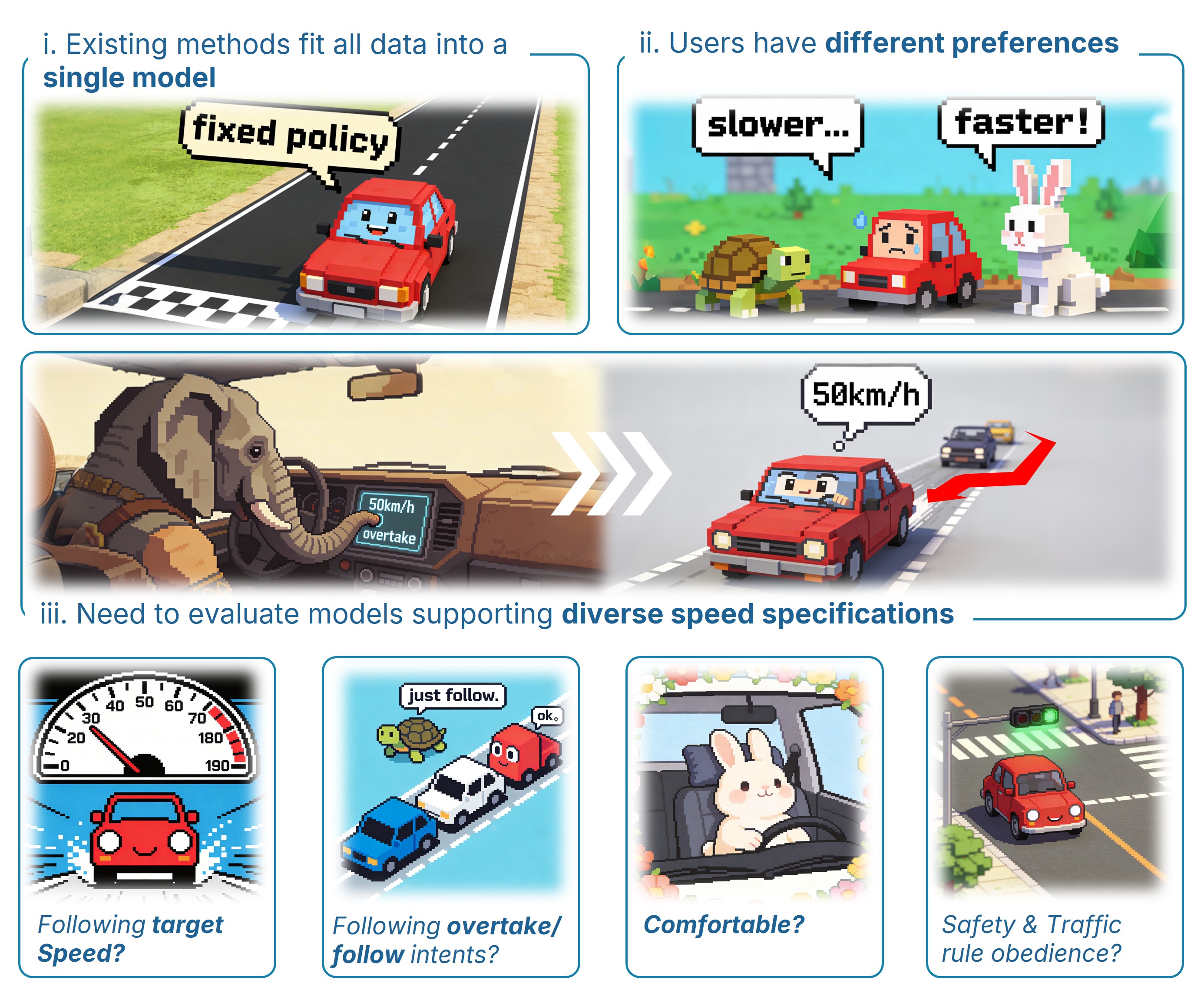}
        \caption{\textbf{Bench2Drive-Speed} introduces target-speed commands and overtake/follow instructions, establishing the first benchmark for desired-speed conditioned autonomous driving. We quantitatively evaluate model controllability across multiple dimensions: speed tracking, interaction strategy, comfort, traffic compliance, \etc.}
        \label{fig:teaser}
        \vspace{-20pt}
    \end{minipage}
    \vspace{-15pt}
\end{figure}

\begin{abstract}
End-to-end autonomous driving (E2E-AD) has achieved remarkable progress. However, one practical and useful function has been long overlooked: users may wish to customize the desired speed of the policy or specify whether to allow the autonomous vehicle to overtake. To bridge this gap, we present \textbf{\ourwork{}}, \textbf{a benchmark with metrics, dataset, and baselines for desired-speed conditioned autonomous driving}. We introduce explicit inputs of users' desired target-speed and overtake/follow instructions to driving policy models. We design quantitative metrics, including Speed-Adherence Score and Overtake Score, to measure how faithfully policies follow user specifications, while remaining compatible with standard autonomous driving metrics. 

To enable training of speed-conditioned policies, one approach is to collect expert demonstrations that strictly follow speed requirements, an expensive and unscalable process in the real world. An alternative is to adapt existing regular driving data by treating the speed observed in future frames as the target speed for training. To investigate this, we construct \textbf{\ourdataset{}}, composed of 2,100 clips annotated with experts demonstrations, enabling systematic investigation of supervision strategies. Our experiments show that, under proper re-annotation, models trained on regular driving data perform comparably to on expert demonstrations, suggesting that \textbf{speed supervision can be introduced without additional complex real-world data collection}. Furthermore, we find that while target-speed following can be achieved without degrading regular driving performance, executing overtaking commands remains challenging due to the inherent difficulty of interactive behaviors. All code,  datasets and baselines are available at \href{https://github.com/Thinklab-SJTU/Bench2Drive-Speed}{https://github.com/Thinklab-SJTU/Bench2Drive-Speed}.

\end{abstract}

\section{Introduction}
\label{sec:intro}

End-to-end autonomous driving (E2E-AD) has achieved significant progress in recent years~\cite{NuScenes, NAVSIM, Bench2Drive}. However, one practical and useful function has long been overlooked: users may wish to customize the desired speed of the driving policy or specify whether to allow the autonomous vehicle to overtake other vehicles. A rushing user may expect the vehicle to maintain higher cruising speeds and overtake slower traffic, while a cautious user may prefer the vehicle to remain behind a leading car even when overtaking is feasible. Despite its clear practical value, this capability remains absent from most existing end-to-end autonomous driving methods and benchmarks.

Ensuring reliable compliance with user-specified speed preferences is challenging. First, in the regular AD data collection process, there is no annotation for the desired target speed. Second, pursuing users' target speeds should not conflict with safety margins, which requires the policies to decide when to follow users' desired speed. Third, achieving a desired speed often requires long-horizon interaction planning, such as overtaking~\cite{tian2022personalized, zhang2023learn}. Fourth, increasing responsiveness to user commands creates inherent trade-offs with comfort~\cite{li2022tradeoff, aledhari2023motion}, as abrupt accelerations or frequent lane changes may compromise passenger experience. 

While traditional planning-and-control (PnC) pipelines can explicitly enforce speed following through structured optimization, such guarantees do not naturally extend to modern E2E-AD systems, where speed behavior emerges implicitly from data-driven policies. Among existing works, style-aware and personalized autonomous driving is most closely related to desired-speed adherence. Early works focused on characterizing human driving styles~\cite{hasenjager2017personalization, liao2024review}, with recent approaches incorporating personalized features into learning-based systems~\cite{schrum2024maveric, kou2025padriver} or leveraging LLM/VLM-based language control~\cite{cui2024onboard, han2024words}. However, in these studies, driving speed is typically embedded within abstract style categories (e.g., Conservative, Normal, Aggressive~\cite{hao2025styledrive}) rather than formulated as an explicit and independently controllable objective. To our knowledge, no existing benchmark provides a principled framework for quantitatively evaluating adherence to user-specified speed commands.

\begin{figure}[t]
\centering
\includegraphics[width=1.0\linewidth]{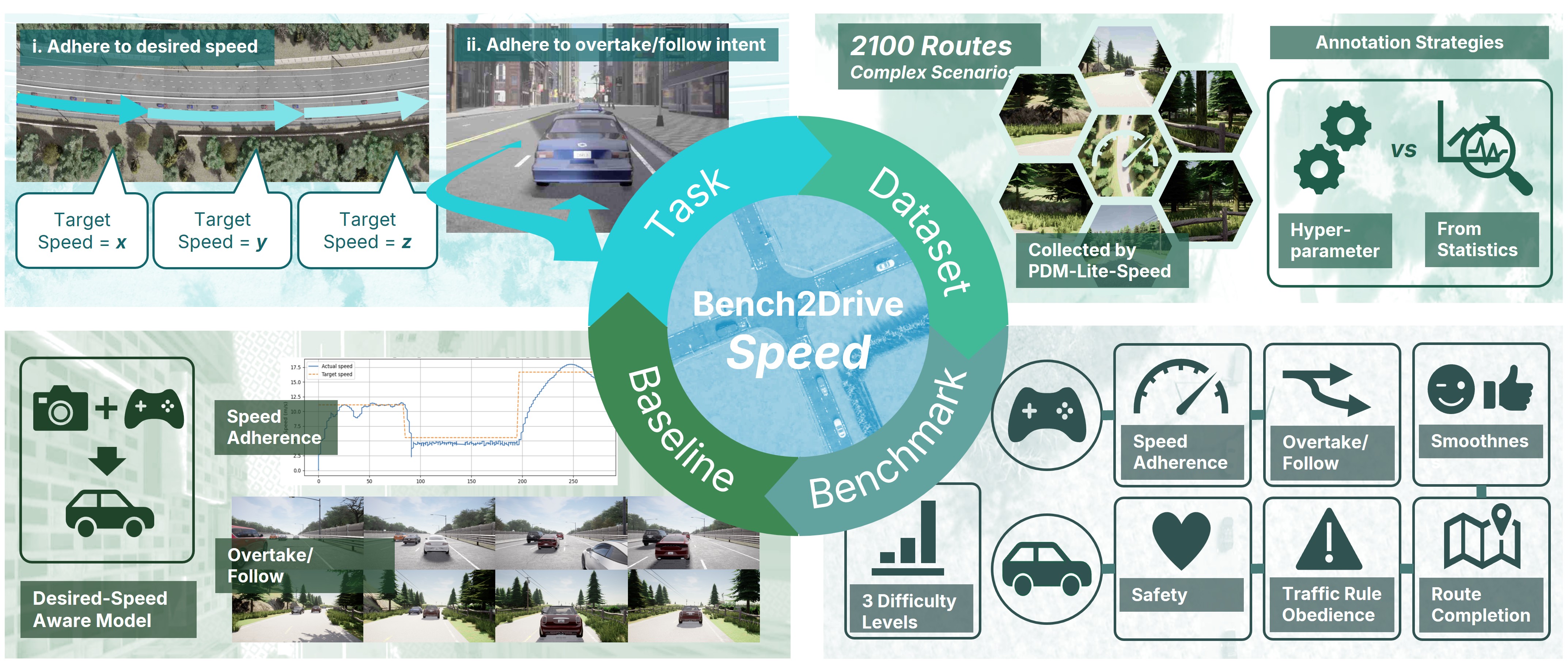}
\caption{We present \textbf{\ourwork{}}, including desired-speed conditioned \textbf{task} - with target speed and overtake/follow commands for speed control; \textbf{dataset} — 2,100 scenarios with extra commands annotated by expert demonstration and virtual target speed strategies; \textbf{benchmark} — controllability metrics (speed adherence, overtake/follow) jointly evaluated with safety, comfort, traffic compliance, \etc; and \textbf{baseline} — the model takes visual and speed command inputs, capable of following target speed commands while attempting to execute overtake/follow behaviors.}
\vspace{-20pt}
\label{fig:overview}
\end{figure}

In this work, we present \textbf{\ourwork{}}, which includes a benchmark, datasets, and baselines for \textbf{desired-speed conditioned autonomous driving}.

\parhead{Benchmark.}
We propose a closed-loop benchmark with the following features:

\begin{enumerate}[leftmargin=10pt, topsep=0pt, itemsep=1pt, partopsep=1pt, parsep=1pt,label=$\bullet$]
\item \textbf{Explicit speed-oriented command.} We introduce \textbf{users' target speed} and \textbf{overtake/follow commands} as an extra input information.

\item \textbf{Quantitative  metrics.} We design \textbf{speed-adherence score} and \textbf{overtake score} to measure how faithfully policies follow given commands.

\item \textbf{Joint evaluation with safety and comfort.} In addition to instruction adherence, the benchmark also reports scores of safety, traffic rule compliance, task completion, and comfort metrics. This enables systematic analysis of how speed controllability trades off with safety and passenger experience.

\item \textbf{Compatibility with standard AD benchmarks.} Policies evaluated under \ourwork{} can also be directly assessed using the Bench2Drive benchmark, ensuring fair comparison with conventional autonomous driving methods within a unified evaluation ecosystem.
\end{enumerate}

\parhead{Dataset.}
We construct \textbf{\ourdataset{}}, consisting of 2,100 complex scenarios collected in CARLA simulator using a modified expert model from~\cite{beisswenger2024pdmlite}, \ouragent{}. Each scenario is annotated with explicit target speed and overtake/follow commands. Importantly, beyond experts' demonstration, which is not in real-world datasets, we further introduce a re-annotation strategy that derives the target speed from the speed of future frames in regular driving data. We refer to this inferred signal as the \textbf{virtual target speed}.

\parhead{Baselines.}
We implement \textbf{\ourbaseline{}} with users' speed command as inputs, based on TCP~\cite{TCP}, under multiple configurations within the proposed benchmark, reporting comprehensive evaluation results. Our experiments demonstrate that models trained with the virtual target speed achieve performance comparable to those trained with target speed from expert demonstration. This finding suggests that reliable target-speed supervision can be introduced in real-world datasets where internal planner parameters are inaccessible.

Notably, while the baselines are able to partially follow the users' target speed without significantly compromising safety, they still struggle to consistently execute overtaking commands. These results reveal a persistent gap between conventional driving performance and explicit speed controllability.

All code, including the evaluation framework, data collection scripts, baselines, and datasets, are publicly released at \href{https://github.com/Thinklab-SJTU/Bench2Drive-Speed}{https://github.com/Thinklab-SJTU/\\Bench2Drive-Speed}.

\section{Related Works}
\label{sec:rel}

\subsection{Benchmarking End-to-End Autonomous Driving}

End-to-end driving models directly map sensory inputs to control signals or trajectories~\cite{codevilla2018end, transfuser, interfuser, UniAD, VAD}. One way to evaluate them is open-loop~\cite{NuScenes}. However, open-loop metrics ignore interactive traffic dynamics and cumulative error~\cite{isEgoStatusAllYouNeed, ADMLP, carla_garage_common_mistakes}. Closed-loop evaluation in simulators such as \carla{}~\cite{CARLA} has therefore become the standard paradigm~\cite{Bench2Drive, CARLALB2, metadrive}. These benchmarks evaluate safety, route completion, \etc under interactive traffic conditions, while semi-closed-loop settings such as NAVSIM~\cite{NAVSIM} enable ego rollouts in non-interactive environments.


\subsection{Style-aware Autonomous Driving}

Among existing works, style-aware and personalized autonomous driving is most closely related to desired-speed adherence. Early works focused on identifying and characterizing human driving styles~\cite{hasenjager2017personalization, liao2024review}, and this line of research continues to grow~\cite{han2024vehicle, yang2024drivingstyle, liu2025diverse}. Some approaches incorporate personalized features into learning-based driving systems~\cite{ling2021towards, zhao2022personalized, natarajan2022adapt, sheng2022study, li2023personalized, ilias2023intelligent, kim2024ndstneuraldrivingstyle, schrum2024maveric}, or leverage them for driver intent prediction~\cite{liao2023driver, jain2016brain4cars, cao2025adapt}. Other methods adapt personalized components to achieve customized performance in specific scenarios~\cite{lu2019personalized, zhu2018personalized, speidel2019towards, sheng2022study, tian2022personalized, chen2024metafollower, cui2024personalized}. Style-aware planners further explore multi-objective trajectory generation that balances safety and efficiency~\cite{li2025learningpersonalized, surmann2025multi, pei2026safe}. More recently, LLM/VLM-based approaches have opened new possibilities~\cite{SimLingo, ORION} by allowing users to change driving styles via language prompts~\cite{cui2024onboard, han2024words, cui2024personalized, kou2025padriver}. 

In several studies~\cite{zhao2022personalized, sheng2022study, surmann2025multi, hao2025styledrive}, driving speed is treated as one aspect of driving style. However, it is typically embedded within abstract style categories (e.g., Conservative, Normal, Aggressive~\cite{hao2025styledrive}) rather than formulated as an explicit and independently controllable objective. Evaluation protocols frequently rely on qualitative case studies~\cite{kou2025padriver, surmann2025multi} or abstract behavioral statistics~\cite{wei2025pdb, hao2025styledrive}.

\begin{table}[t]
\centering
\caption{\textbf{Comparison of Driving Benchmarks}. \ourwork{} is the first benchmark to support closed-loop evaluation, end-to-end learning, and explicit speed-conditioned control for autonomous driving.}
\label{tab:benchmark_comparison}
\setlength{\tabcolsep}{8pt}
\resizebox{\textwidth}{!}{%
\begin{tabular}{lccccc}
\toprule
\textbf{Benchmark} & \textbf{Source} & \textbf{Closed-Loop} & \textbf{E2E} & \textbf{Style-Aware} & \textbf{Speed-Conditioned} \\
\midrule
nuScenes~\cite{NuScenes} & Real & \ding{55} & \ding{51} & \ding{55} & \ding{55} \\
CARLA~\cite{CARLA, CARLALB2} & Simulator & \ding{51} & \ding{51} & \ding{55} & \ding{55} \\
MetaDrive~\cite{metadrive} & Simulator & \ding{51} & \ding{51} & \ding{55} & \ding{55} \\
Bench2Drive~\cite{Bench2Drive} & Simulator & \ding{51} & \ding{51} & \ding{55} & \ding{55} \\
NAVSIM~\cite{NAVSIM} & Real & \ding{55} & \ding{51} & \ding{55} & \ding{55} \\
StyleDrive~\cite{hao2025styledrive} & Real & \ding{55} & \ding{51} & \ding{51} &\ding{55} \\
\midrule
\ourwork{} (Ours) & Simulator & \ding{51} & \ding{51} & \ding{51} & \ding{51} \\
\bottomrule
\end{tabular}
}
\\[3pt]
\vspace{-18pt}
\end{table}

\section{\ourwork}
\label{sec:work}

We present \ourwork{}, which enables systematic evaluation of autonomous driving under explicit target-speed control. In this section, we give the task formulation (Sec. ~\ref{sec:scenes}), describe the construction and annotation pipeline of \ourdataset{} (Sec.~\ref{sec:dataset}), and finally detail the evaluation protocol (Sec.~\ref{sec:eval}).

\subsection{Desired Speed Conditioned Autonomous Driving}
\label{sec:scenes}

Standard input formulations in AD models do not expose explicit control interfaces for desired-speed regulation, nor do existing scenarios evaluate speed adherence. To bridge this gap, we introduce additional users' speed customization as inputs and dedicated scenario designs that explicitly evaluate compliance with target-speed instructions.

\parhead{Extra Command Inputs.} 
We define two high-level control commands: a \textbf{target-speed command} and an \textbf{overtaking command}. 
The target-speed command provides a direct interface for controlling the model’s longitudinal velocity. The overtaking command specifies whether the policy should overtake a slower leading vehicle or remain behind it when traffic conditions permit.

\parhead{Scenario Implementation.}
The closed-loop evaluation and data collection pipeline of \ourwork{} is built upon the \carla{}~\cite{CARLA} simulator, with extensions to support command-conditioned evaluation.

First, we augment route configuration files with segment-wise target-speed specifications. Our evaluation pipeline parses these configurations and supplies target-speed signals according to the ego vehicle’s location.

Second, to systematically evaluate overtaking behavior, we introduce a dedicated scenario containing a slow-moving lead vehicle. 
Under the \texttt{overtake} command, the policy is expected to pass the leading vehicle; under the \texttt{follow} command, it must maintain its position behind the vehicle. 

\begin{figure}[t]
    \centering
    \includegraphics[width=0.9\linewidth]{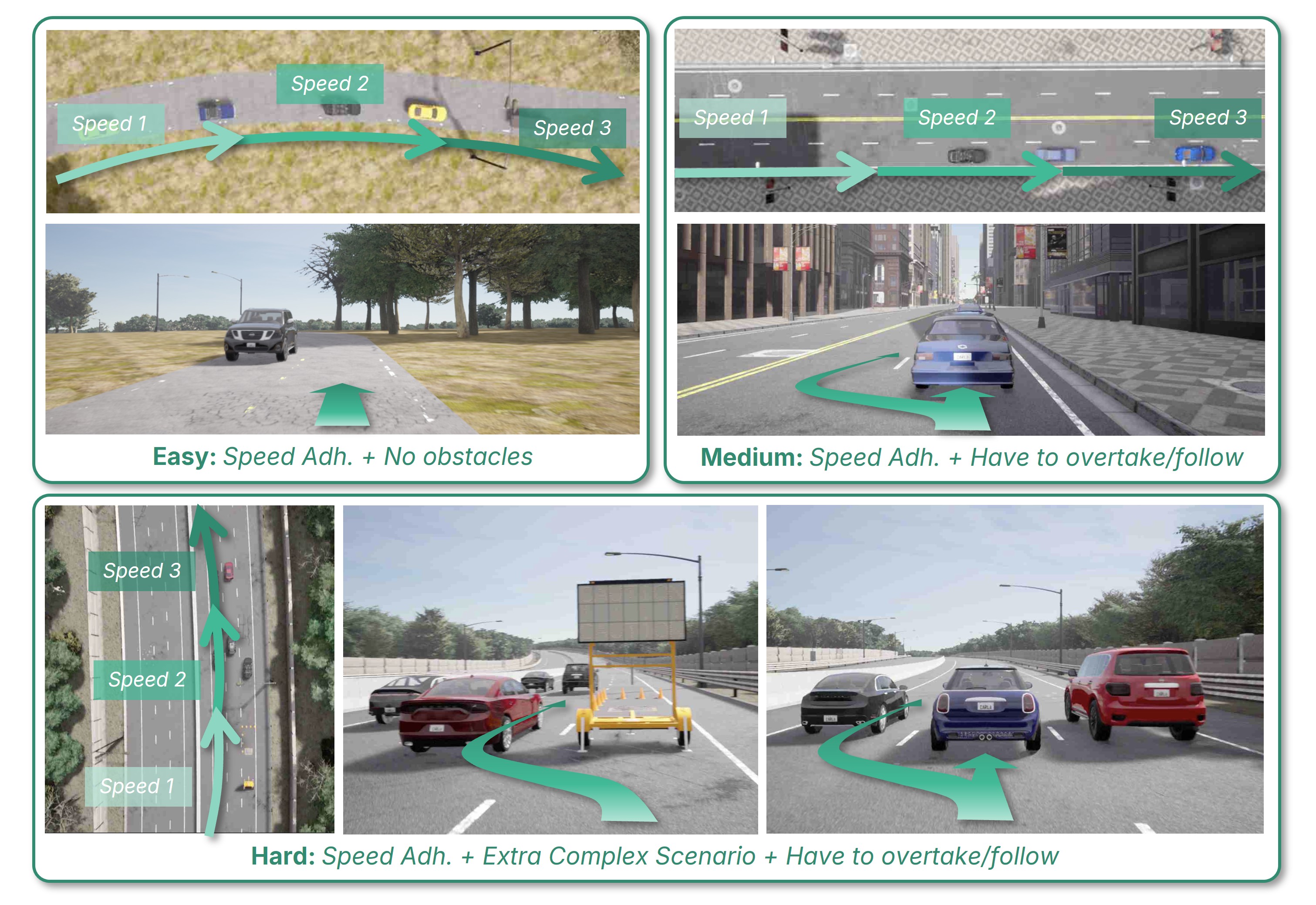}
    \caption{\textbf{Three Difficulty Levels in \ourwork{}}. The difficulty of adhering to the desired speed increases from easy to hard. Overtaking and following adherence are evaluated only in the medium and hard scenarios.}
    \vspace{-15pt}
    \label{fig:difficulty_illustration}
\end{figure}

\parhead{Diverse Difficulty.} 
The dataset and evaluation routes in \ourwork{} are organized into three difficulty levels to enable stratified analysis of controllability performance under progressively more challenging scenarios (Fig.~\ref{fig:difficulty_illustration}):

\textbf{Easy}: Routes without interfering vehicles, allowing pure evaluation of command adherence.

\textbf{Medium}: Routes containing slower vehicles in the ego lane, requiring lane changes and overtaking maneuvers to maintain target-speed compliance.

\textbf{Hard}: Complex traffic scenarios adapted from the \carla{} Leaderboard v2~\cite{CARLA, CARLALB2}, where the ego vehicle must handle dynamic incidents while simultaneously satisfying the given control commands.

\subsection{Dataset}
\label{sec:dataset}

\begin{figure}[t]
    \centering
    \includegraphics[width=1.0\linewidth]{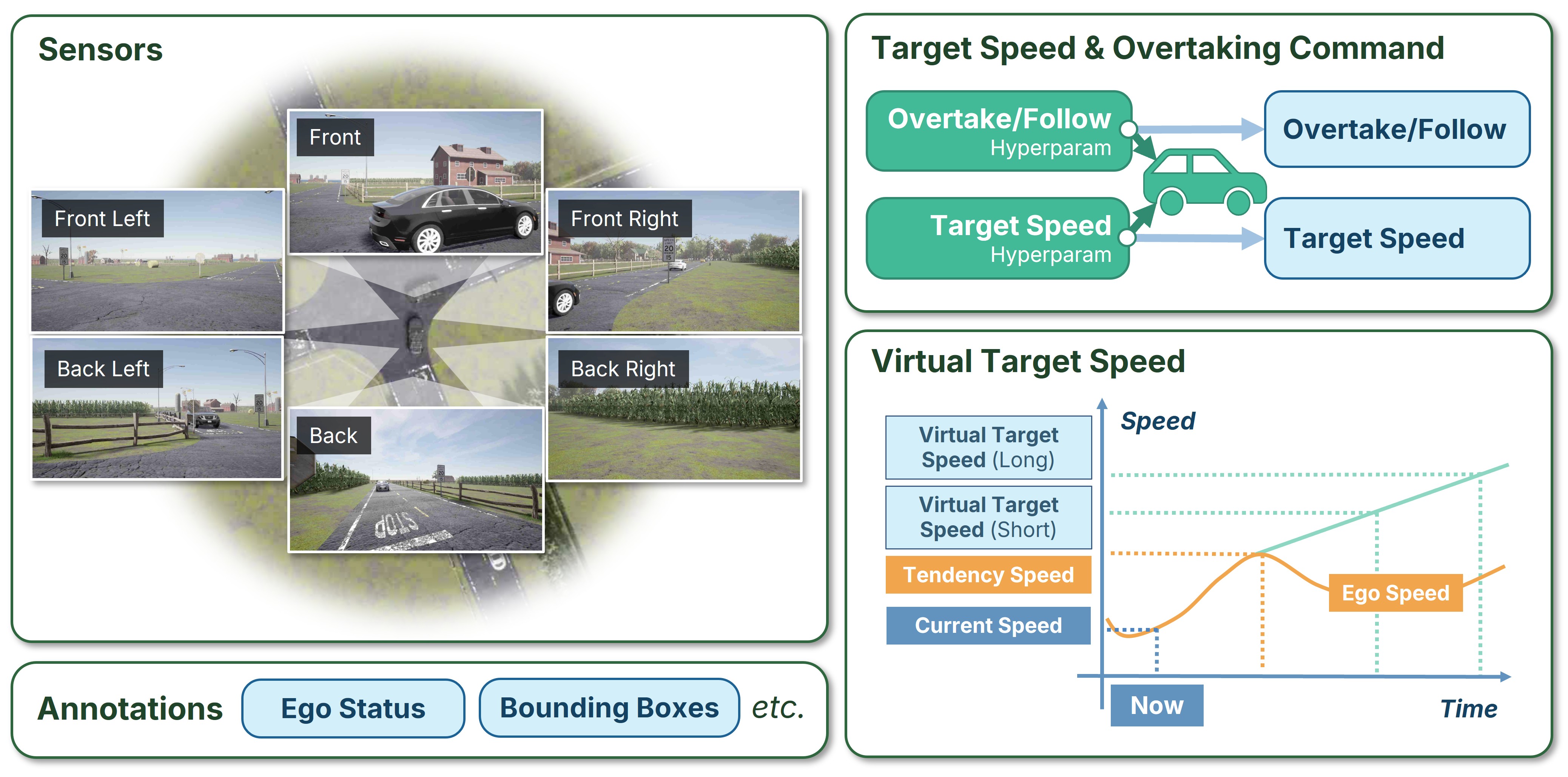}
    \vspace{-18pt}
    \caption{\textbf{Illustration of \ourdataset{}}. The dataset includes visual sensor inputs, ego-state information, bounding box annotations, overtaking commands, and target-speed commands from different sources.}
    \vspace{-12pt}
    \label{fig:data_layout}
\end{figure}

\ourdataset{} consists of 2,100 driving scenes collected in the CARLA simulator by modified rule-based expert model \ouragent{}. The dataset provides the same data format as Bench2Drive~\cite{Bench2Drive}, as well as newly introduced speed commands, including target speed and overtaking instructions, as shown in Fig.~\ref{fig:data_layout}.

\begin{figure}[t]
    \centering
    \includegraphics[width=1.0\linewidth]{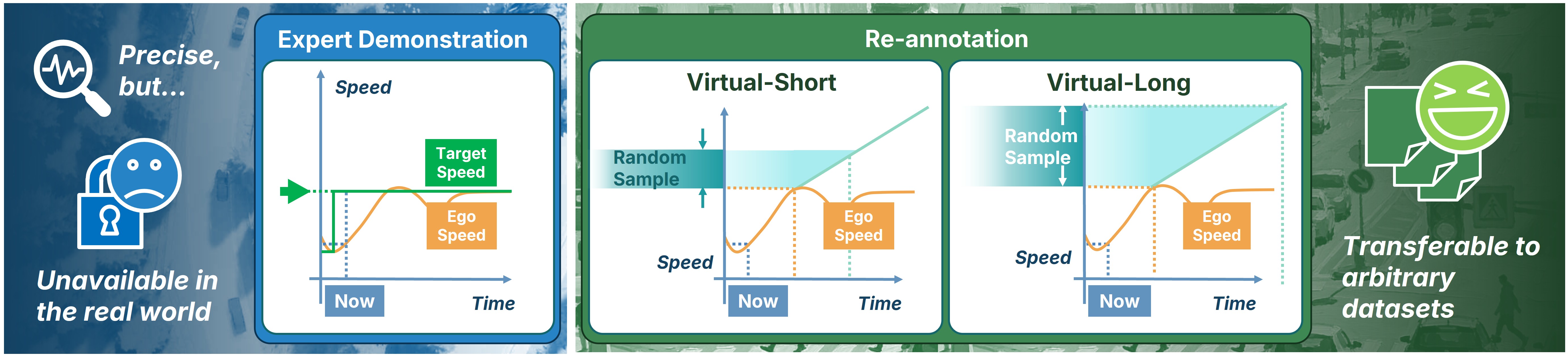}
    \vspace{-18pt}
    \caption{\textbf{Illustration of different target-speed annotation methods.} Expert demonstrations are precise but rely on the data collection model's internal hyperparameters—which are unavailable in practice—whereas re-annotation is more feasible.}
    \vspace{-12pt}
    \label{fig:speed_anno}
\end{figure}

\parhead{Data Collection.} \ouragent{} is implemented based on PDM-Lite~\cite{beisswenger2024pdmlite, DriveLM}, which is inspired by PDM-Closed~\cite{dauner23aPDM-Closed} and IDM~\cite{treiber2000idm}. It leverages privileged information provided by the \carla{} simulator, and demonstrates strong performance in challenging scenarios from \carla{} Leaderboard 2.0. To support user-controllable speed-conditioned driving, \ouragent{} is extended with explicit mechanisms for following speed-related commands and enhanced to handle the newly introduced scenarios in Sec.~\ref{sec:scenes}.

\parhead{Target Speed Annotation.}\label{sec:target_speed}
While overtake/follow commands can be directly derived from expert behaviors, defining a desired target speed is less straightforward. We consider two supervision strategies for further discussions.

\begin{enumerate}[leftmargin=10pt, topsep=0pt, itemsep=1pt, partopsep=1pt, parsep=1pt]
    \item \textbf{Expert Demonstration.} Since \ourdataset{} is collected using \ouragent{}, the target speed can be obtained from its internal cruising-speed hyperparameter, reflecting the intended velocity under hazard-free conditions. This supervision signal is precise but relies on privileged access to the expert controller and is therefore \textbf{unavailable in regular real-world datasets, which makes it a strong yet impractical baseline}.
    
    \item \textbf{Re-annotation Regular Driving Data.}
    To avoid dependence on internal controller parameters, we introduce a re-annotation strategy, referred to as \textbf{Virtual Target Speed}. It is computed in two stages. First, a \emph{tendency speed} is extracted from short-horizon future speed sequences by identifying the maximal monotonic speed change trend. \textbf{Secondly, to avoid information leakage, this tendency is extrapolated over a randomized temporal window}, as shown in Fig.~\ref{fig:speed_anno}.

    

    To analyze the influence of extrapolation strength, we provide two configurations for subsequent experiments:

    \begin{itemize}
        \item \textbf{Long:} longer temporal horizon and larger extension bound;
        \item \textbf{Short:} shorter horizon and conservative extension.
    \end{itemize}

    This re-annotation strategy does not require privileged planner parameters, enabling the direct usage of existing massive driving data.
\end{enumerate}

\begin{wraptable}{r}{0.5\linewidth}
\vspace{-32pt}
\centering
\small
\setlength{\tabcolsep}{6pt}
\caption{\textbf{Composition of \ourdataset{} across difficulty levels and overtake/follow options.}}
\label{tab:dataset_split}
\begin{tabular}{lcc}
\toprule
\textbf{Difficulty} & \textbf{Overtake} & \textbf{Follow} \\
\midrule
Medium & 570 & 570 \\
Hard   & 480 & 480 \\
\midrule
\textbf{Total} & \textbf{1,050} & \textbf{1,050} \\
\bottomrule
\end{tabular}
\vspace{-24pt}
\end{wraptable}
\parhead{\ourdataset{}} consists of 2,100 driving routes collected in \carla{}. The dataset is organized along two axes: difficulty (medium, hard) and overtaking decision (overtake, follow), resulting in four domains. The detailed distribution is summarized in Table~\ref{tab:dataset_split}. Each domain contains a comparable number of routes to maintain a balanced distribution across difficulty and behavior settings. \ourdataset{} has the following features:

\begin{figure}[t]
    \centering
    \includegraphics[width=1.0\linewidth]{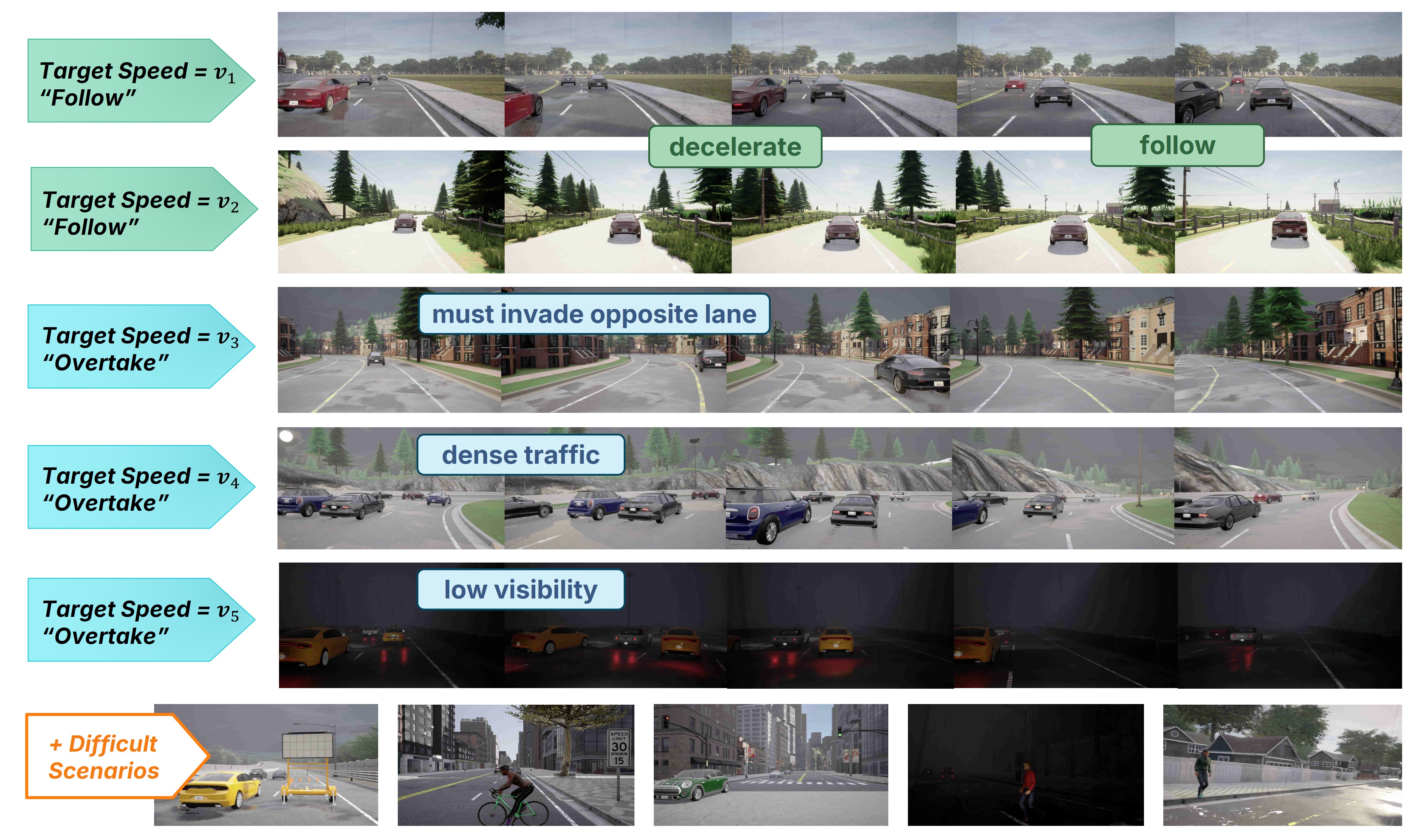}
    \vspace{-18pt}
    \caption{\textbf{Diverse scenarios in \ourdataset{}.} The ego vehicle adheres to target-speed and overtaking commands across various situations, with ample coverage of complex scenarios.}
    \vspace{-12pt}
    \label{fig:first_person}
\end{figure}

\begin{figure}[t]
    \centering
    \begin{subfigure}[b]{0.36\linewidth}
        \centering
        \includegraphics[width=\linewidth]{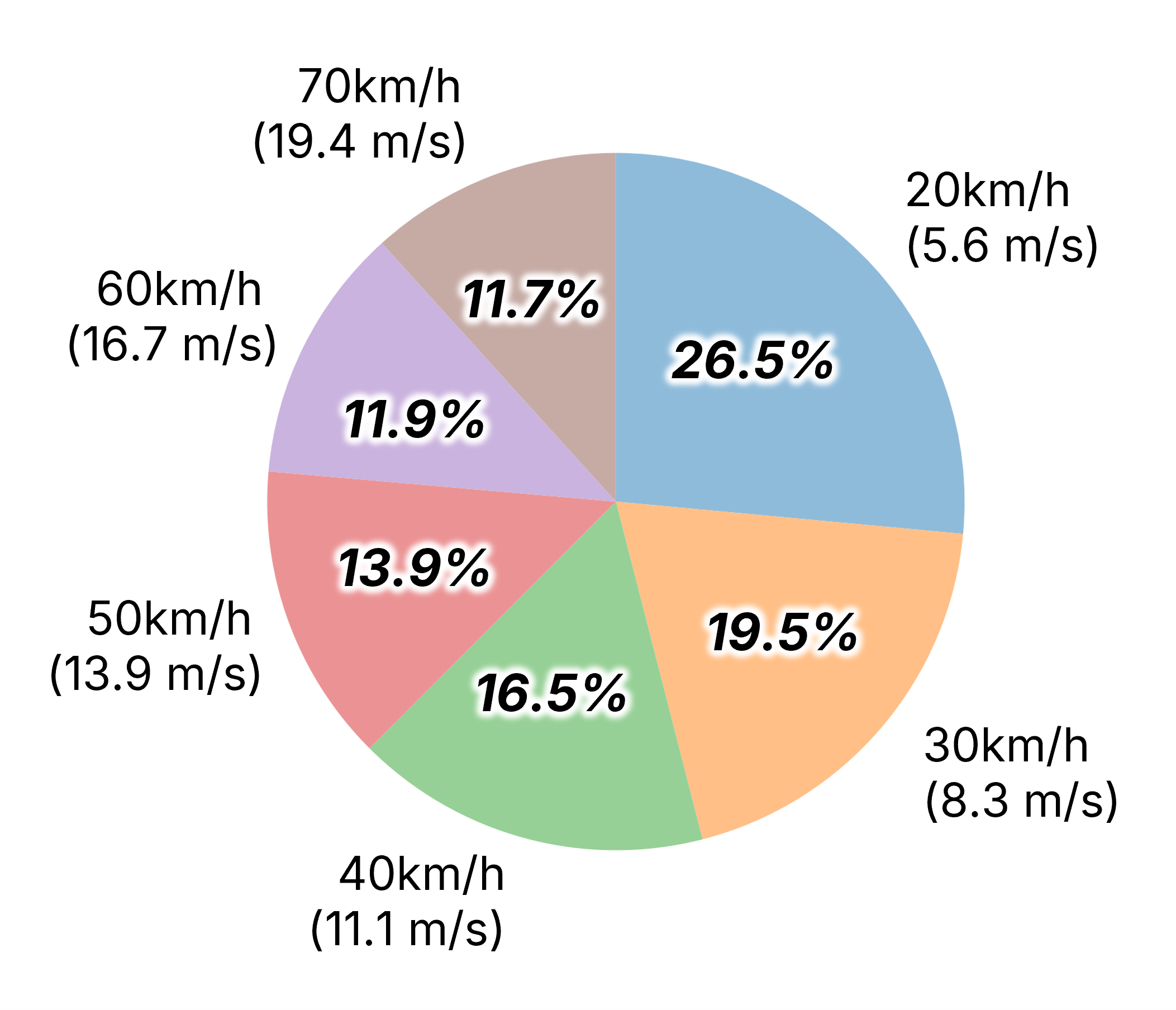}
        \label{fig:target_speed_distribution}
    \end{subfigure}
    \hspace{0.03\linewidth}
    \begin{subfigure}[b]{0.54\linewidth}
        \centering
        \includegraphics[width=\linewidth]{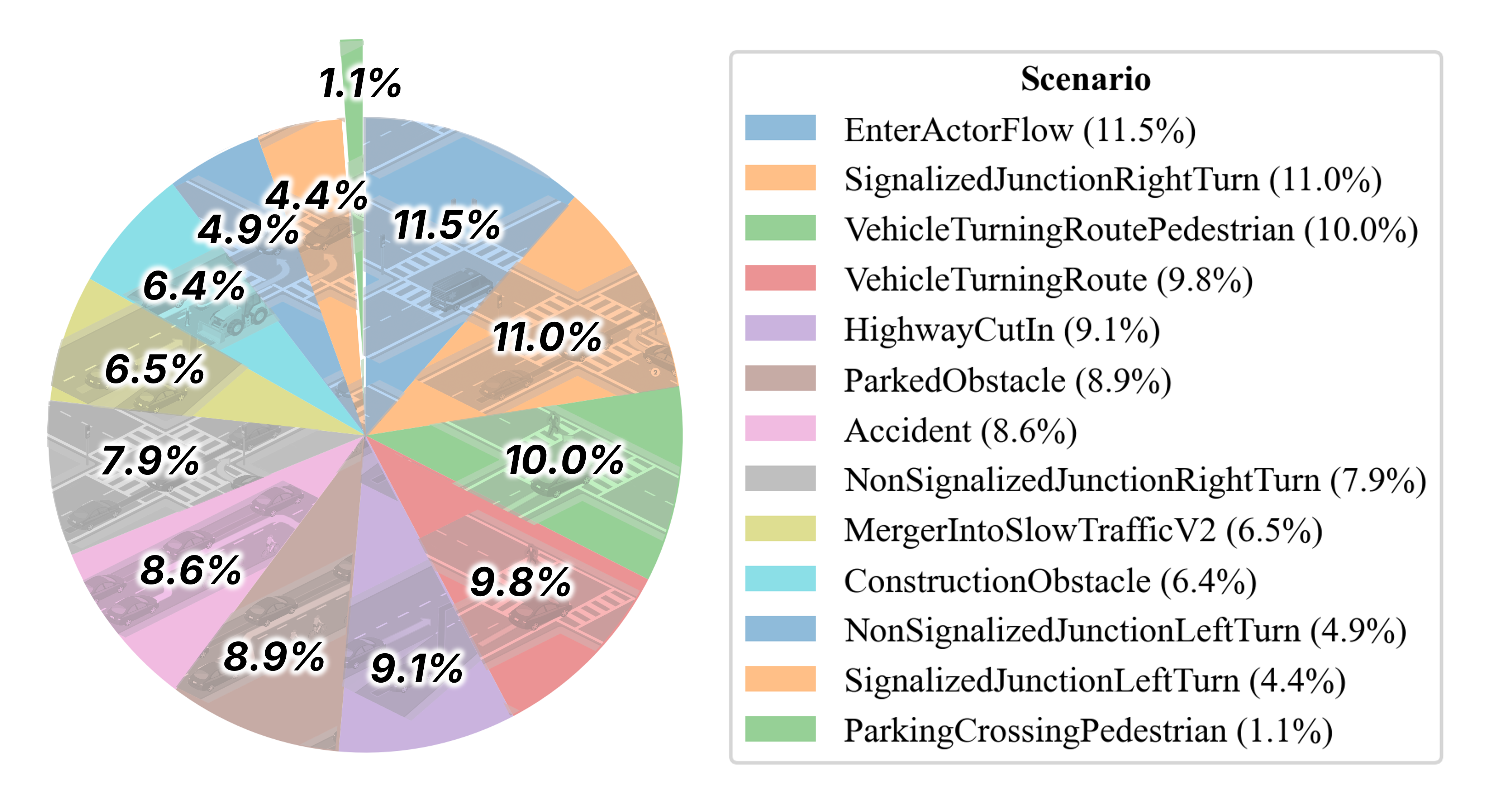}
        \label{fig:scenario_distribution}
    \end{subfigure}
    \vspace{-6mm}
    \caption{\textbf{Target speed distribution} in \ourdataset{} (Left) and \textbf{Distribution of difficult scenarios} included in \ourdataset{}, following CARLA Leaderboard v2~\cite{CARLALB2} (Right).}
    \vspace{-3mm}
    \label{fig:balanced_speed_scenario}
\end{figure}

\begin{itemize}[leftmargin=10pt, topsep=0pt, itemsep=1pt, partopsep=1pt, parsep=1pt,label=$\bullet$]
    \item \textbf{Challenging and Diverse Scenarios.}
    To expose the model to complex and safety-critical traffic conditions during training, the hard routes in \ourdataset{} incorporate 13 of the most demanding scenarios from the \carla{} Leaderboard v2~\cite{CARLALB2} (Fig.~\ref{fig:balanced_speed_scenario}), including static obstacle circumvention, dense traffic merging, multi-directional junction negotiation, and yielding to pedestrians and cyclists. By integrating these challenging situations, \ourdataset{} provides diverse supervisory trajectories that jointly stress safety, task completion, and desired-speed adherence (Fig.~\ref{fig:first_person}).
    
    \item \textbf{Balanced Distribution Across Speed Commands and Environment Factors.} To ensure unbiased supervision, \ourdataset{} enforces a balanced distribution across key controllability and environmental dimensions, including target-speed commands(Fig.~\ref{fig:balanced_speed_scenario}), difficult scenarios, and weather conditions, \etc.
    
    \item \textbf{Within-Route Command Variability.}
    Unlike conventional datasets where control characteristics remain largely stationary within a route, \ourdataset{} assigns varying target-speed commands across different route segments. This design requires the ego vehicle to adapt its longitudinal behavior dynamically during execution, rather than relying on persistent temporal patterns, mitigating shortcut learning and reducing potential causal leakage.
\end{itemize}

\subsection{Evaluation}
\label{sec:eval}

In this section, we first describe the setup of our closed-loop evaluation scenarios, followed by the detailed metrics used for assessment.

\parhead{Scenario Settings.}
To construct a comprehensive closed-loop evaluation set, we select 16 representative scenarios for each of the three difficulty levels, resulting in a total of 48 evaluation cases. Each case consists of four distinct routes paired with four different sets of speed commands. This design \textbf{prevents shortcut learning based solely on specific road layouts or fixed traffic situations}, and ensures that model behavior differences arise from command execution rather than memorization of scenes. The detailed composition of the evaluation scenarios is summarized in Table~\ref{tab:scenario_settings}.

\begin{table}[t]
    \centering
    \setlength{\tabcolsep}{6pt}
    \vspace{-0mm}\caption{\textbf{Scenarios of  Different Difficulty Levels}.}
    \label{tab:scenario_settings}
    \resizebox{\textwidth}{!}{%
    \begin{tabular}{lll}
        \toprule
        \textbf{Difficulty} & \textbf{Route Layout} & \textbf{Special Incidents} \\
        \midrule
        Easy & Rural curving road & None \\
        Easy & Left turn at urban intersection & None \\
        Easy & Straight through urban intersection & None \\
        Easy & Wide street & None \\
        \midrule
        Medium & Rural curving road & \makecell[l]{Overtake or follow one vehicle} \\
        Medium & Wide street & \makecell[l]{Overtake or follow one vehicle} \\
        Medium & Right turn at rural intersection & \makecell[l]{Overtake or follow one vehicle} \\
        Medium & Straight through urban intersection & \makecell[l]{Overtake or follow one vehicle} \\
        \midrule
        Hard & Wide street & \makecell[l]{Accident avoidance + overtake or follow one vehicle} \\
        Hard & Wide street & \makecell[l]{Construction obstacle avoidance + overtake or follow one vehicle} \\
        Hard & Right turn at rural intersection & \makecell[l]{Junction handling + overtake or follow one vehicle} \\
        Hard & Left turn at urban intersection & \makecell[l]{Junction handling + pedestrian yielding + \\ overtake or follow one vehicle} \\
        \bottomrule
    \end{tabular}%
    }
    \vspace{-12pt}
\end{table}

\parhead{Metrics.} Our evaluation metrics are designed to quantify command adherence while preserving compatibility with standard autonomous driving benchmarks, which encompass the following metrics: 



\textbf{Speed-Adherence Score.}
We evaluate whether a policy adjusts its longitudinal behavior according to the user-specified target speed in closed-loop driving. Given a reference route $\mathcal{R}$ of length $L$ and a closed-loop trajectory
$\{(x_i, y_i, \mathbf{v}_i)\}_{i=1}^N$,
we project each ego position onto the route to obtain its arc-length coordinate $s_i \in [0,L]$.
The actual speed is computed as $v^{\text{actual}}_i=\|\mathbf{v}_i\|_2$,
and the target speed is from the user-defined speed profile
$v^{\text{target}}_i = v^{\text{target}}(s_i)$.

To reduce bias from non-uniform sampling or stationary states, we adopt distance-based weighting with
$w_i = \|(x_i,y_i)-(x_{i-1},y_{i-1})\|_2$.
The relative speed error is defined as
\begin{equation}
e_i = \frac{|v^{\text{actual}}_i - v^{\text{target}}_i|}
{\max(v^{\text{target}}_i, \epsilon)} ,
\end{equation}
and converted into a per-step score $\text{score}_i = \exp(-\alpha e_i)$, where $\alpha$ controls the penalty strength.

The overall compliance score is computed as a distance-weighted average:
\begin{equation}
\text{Score}_{\text{speed}} =
\frac{\sum_{i=2}^{N} w_i \cdot \text{score}_i}
{\sum_{i=2}^{N} w_i}.
\end{equation}

Especially, in \textit{Follow} scenarios, when the ego vehicle is constrained by a slower lead vehicle
($v^{\text{lead}} \le v^{\text{actual}}_i < v^{\text{target}}_i$), we soften the penalty.

\textbf{Overtake Score.} For each route, all scenarios with explicit \textit{overtake} or \textit{follow} commands are evaluated. A scenario is considered successful only if it is properly triggered and the model executes the commanded behavior correctly. Each scenario receives a binary score (100 or 0), and the final route-level score is computed as the success ratio over all required scenarios. Scenarios that fail to activate (e.g., due to not reaching the trigger waypoint) are counted as failures \textbf{to prevent models from artificially inflating performance by avoiding difficult scenarios}.

\textbf{Comfort, Safety, and Route Completion Score.}
As our framework is built upon Bench2Drive~\cite{Bench2Drive}, all standard \carla{} and Bench2Drive evaluation metrics are fully supported. 
These include Driving Score (DS), Success Rate (SR), multi-ability metrics, efficiency, and comfort. 

\section{Experiments}
\label{sec:expr}

In this section, we conduct comprehensive closed-loop evaluations to assess speed-conditioned driving performance under varying traffic complexity and command settings. We first describe the implemented baselines and the constitution of training datasets in Sec.~\ref{sec:baselines}. We then present quantitative results and detailed analyses in Sec.~\ref{sec:results}.

\subsection{Baselines and Datasets}
\label{sec:baselines}

\begin{figure}[t]
\centering
\includegraphics[width=1.0\linewidth]{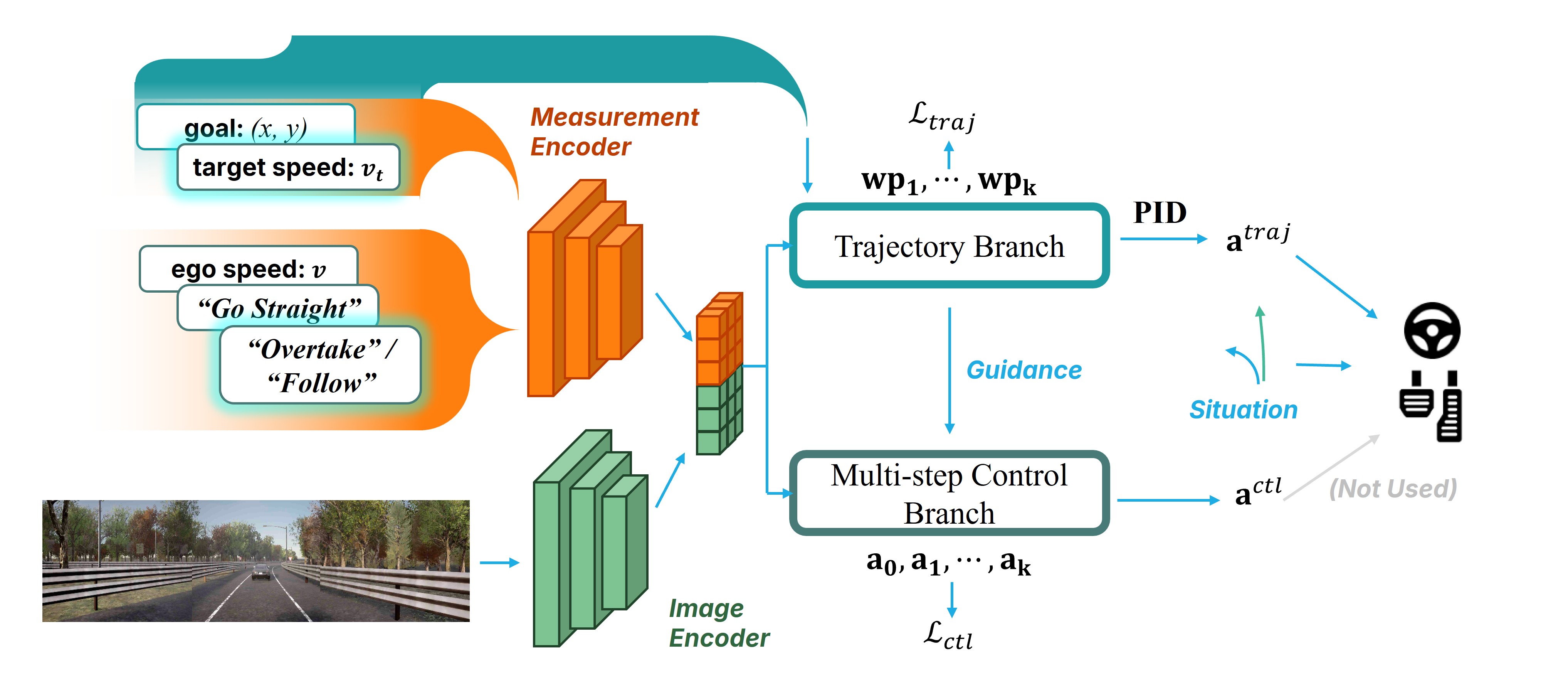}
\caption{\textbf{Overview of \ourbaseline{}.} The encoded ego state and driving commands are concatenated with features extracted from RGB inputs. The fused representation is then fed into both the trajectory and control branches. Additionally, the target waypoint(goal) and target-speed command are used to guide trajectory generation. In practice, only the output of the trajectory branch is utilized.}
\vspace{-10pt}
\label{fig:baseline}
\end{figure}

\parhead{Baselines.} Our implemented baseline, \ourbaseline{}, is built upon TCP\cite{TCP}, due to its simplicity. Note that \textbf{our speed-command-related designs are model-agnostic.} For \ourbaseline{}, to align with the Bench2Drive setup, we adopt TCP's trajectory-only variant for closed-loop implementation. In \ourbaseline{}, the target speed and overtake command are concatenated into the model input to enable speed-conditioned driving, as shown in Fig.~\ref{fig:baseline}.

\parhead{Training Dataset Design.} 
To study the effect of different supervision sources, we construct multiple dataset variants by combining data sources and target-speed annotation strategies. Bench2Drive-base1000~\cite{Bench2Drive} contains 1000 scenes without speed commands, which only supports re-annotation of virtual speed. Our extended dataset records the desired speed, while it could be re-annotated with virtual target speed as well. The resulting dataset variants are summarized in Table~\ref{tab:dataset_variants}.

\begin{table}[!t]
\centering
\setlength{\tabcolsep}{4pt}
\caption{\textbf{Different Training Data Configurations}.}
\vspace{-4mm}
\resizebox{\linewidth}{!}{%
\begin{tabular}{l|c|cc}
\toprule
\multirow{2}{*}{\textbf{Data Source}} 
& \textbf{From Expert} 
& \multicolumn{2}{c}{\textbf{From Re-annotation (Virtual Target Speed)}} \\
\cline{3-4}
& (Hyper-parameter) 
& Long Extension 
& Short Extension \\
\midrule
\btd{}
& -- 
& \btdlong{} 
& \btdshort{} \\
\ourdataset{} 
& \expertdata{}
& \virtuallong{}
& \virtualshort{} \\
\bottomrule
\end{tabular}%
}
\vspace{-4mm}
\label{tab:dataset_variants}
\end{table}


\subsection{Results.}
\label{sec:results}

\begin{figure}[t]
\centering
\includegraphics[width=1.0\linewidth]{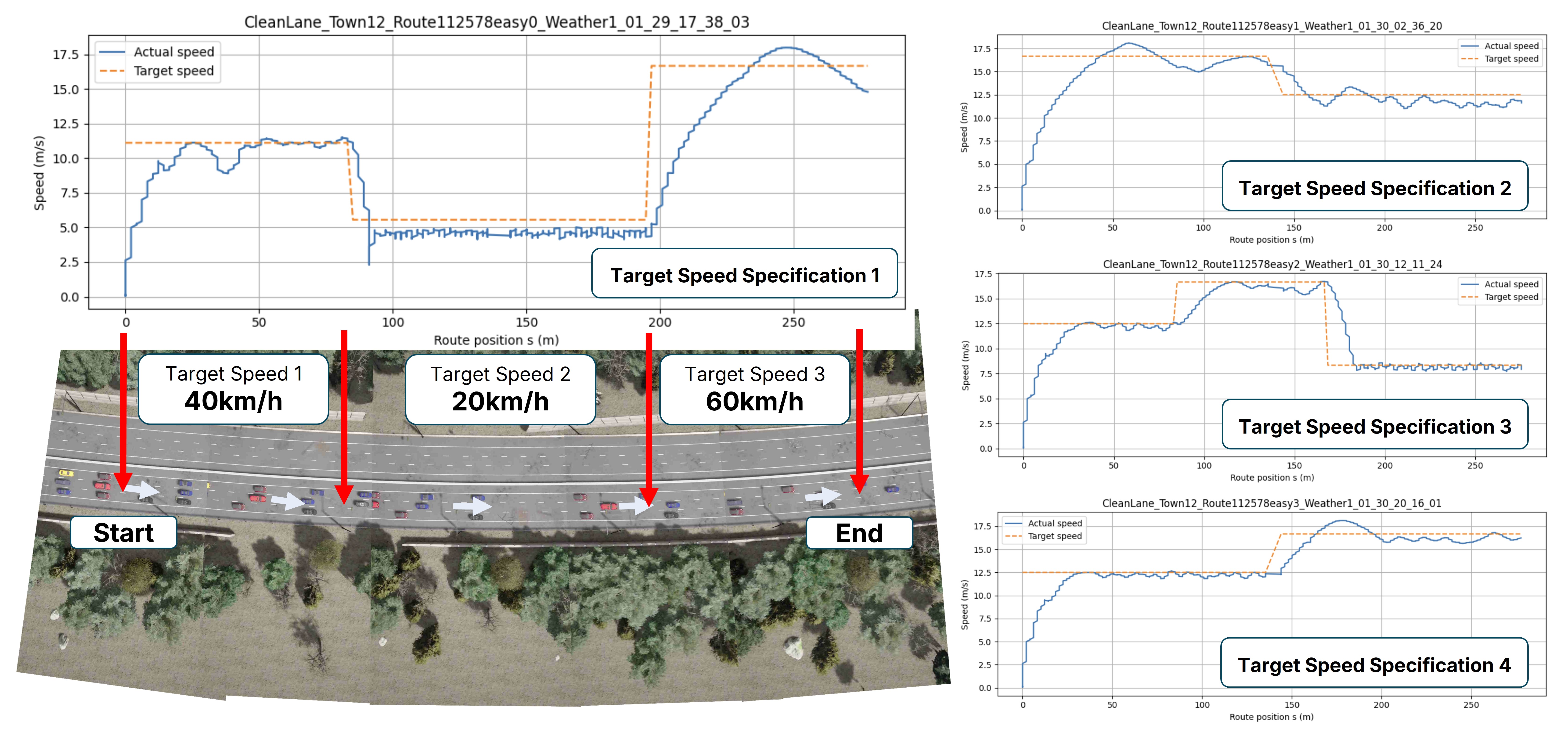}
\caption{\textbf{Visualization of Speed Profile}. The target speed adherence of \ourbaseline{} under the same route with different target speed specifications.}
\vspace{-15pt}
\label{fig:target_speed}
\end{figure}

\begin{figure}[t]
\centering
\includegraphics[width=0.75\linewidth]{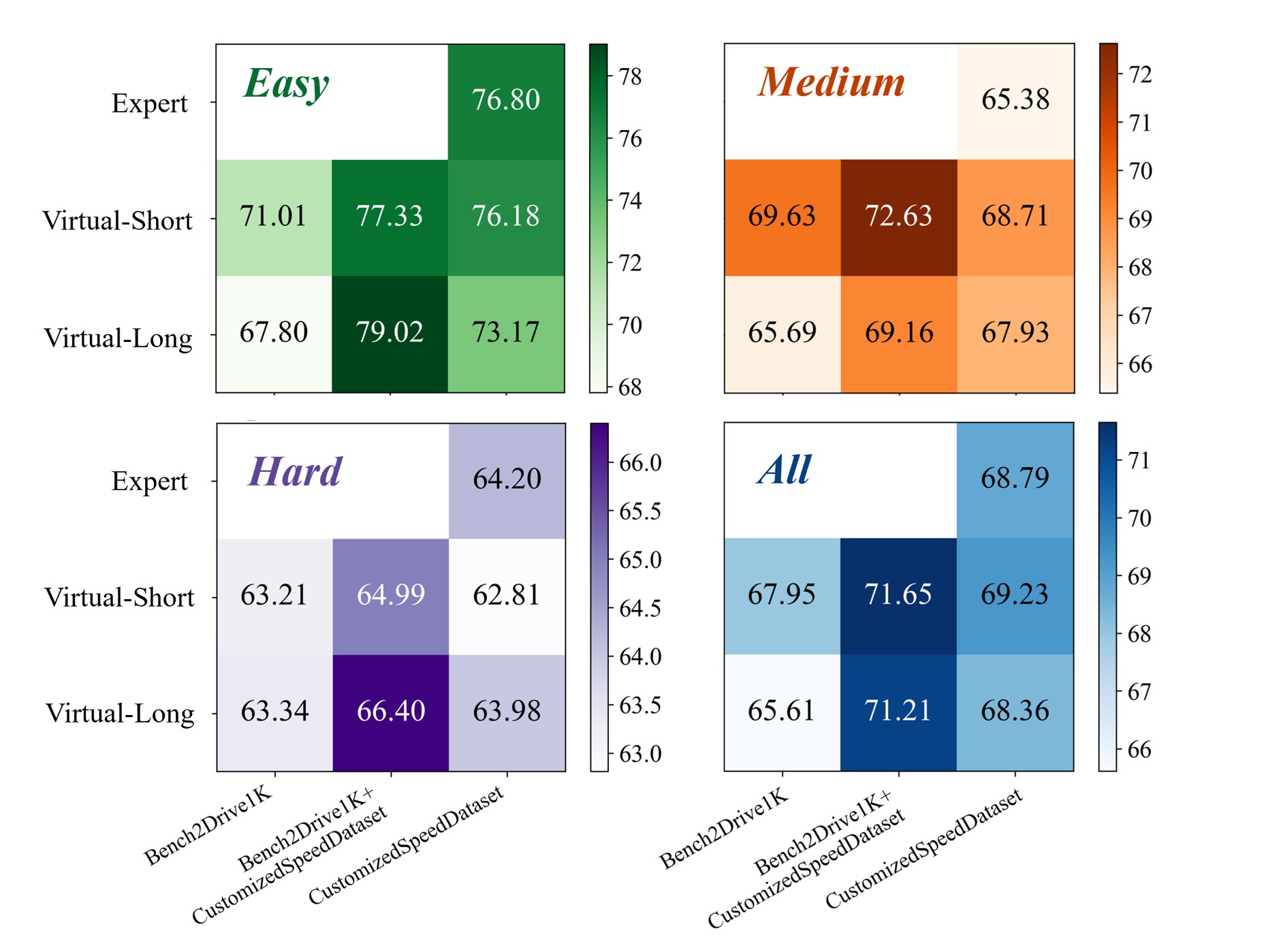}
\caption{\textbf{Heatmap of \ourbaseline{} speed-adherence scores for models trained on different datasets, evaluated on Easy, Medium, Hard, and All routes.} Models trained with virtual target-speed supervision achieve performance comparable to those trained on expert demonstrations. Speed adherence is generally higher under the virtual-short setting than virtual-long.}
\vspace{-3mm}
\label{fig:heatmap_speed}
\end{figure}

\begin{table*}[t]
\centering
\setlength{\tabcolsep}{5pt}
\caption{\textbf{Speed-Adherence Score and Overtake Score} on 48 evaluation routes of \ourwork{}. Metrics are reported for All(A), Easy (E), Medium (M), and Hard (H). The best result of each score is highlighted in bold.}
\label{tab:speed_overtake}
\resizebox{1.0\textwidth}{!}{
\begin{tabular}{ll|cccc|cccc}
\toprule
\multirow{2}{*}{\textbf{Model}} 
& \multirow{2}{*}{\textbf{Dataset}}
& \multicolumn{4}{c|}{\textbf{Speed-Adherence Score $\uparrow$}}
& \multicolumn{4}{c}{\textbf{Overtake Score $\uparrow$}} \\
\cmidrule{3-6}
\cmidrule{7-10}
& 
& \cellcolor{gray!15}\textbf{A} & E & M & H
& \cellcolor{gray!15}\textbf{A} & E & M & H \\
\midrule

$\text{TCP}_{\text{w/o Speed Command}}$ & \ourdataset{}
& \cellcolor{gray!15}41.54 & 42.00 & 40.95 & 41.67
& \cellcolor{gray!15}18.75 & - & 37.50 & 0.00 \\

TCP-Speed & \expertdata{}
& \cellcolor{gray!15}68.79 & \textbf{76.80} & 65.38 & \textbf{64.20}
& \cellcolor{gray!15}21.88 & - & 37.50 & 6.25 \\

TCP-Speed & \virtualshort{}
& \cellcolor{gray!15}\textbf{69.23} & 76.18 & \textbf{68.71} & 62.81
& \cellcolor{gray!15}\textbf{40.63} & - & \textbf{56.25} & \textbf{25.00} \\

TCP-Speed & \virtuallong{}
& \cellcolor{gray!15}68.36 & 73.17 & 67.93 & 63.98
& \cellcolor{gray!15}\textbf{40.63} & - & \textbf{56.25} & \textbf{25.00} \\

\bottomrule
\end{tabular}
}
\vspace{-8mm}
\end{table*}

We evaluate models trained from scratch on the 48 routes of \ourwork{} under \textbf{different training dataset combinations to analyze the impact of different target speed annotation strategies}. Additionally, we assess their performance on Bench2Drive-220~\cite{Bench2Drive} routes with a default command specification of \emph{30 km/h, follow} to assess whether traditional closed-loop driving abilities are preserved. Based on the evaluation results, the following questions could be answered:

\parhead{1. To what extent can the baseline follow target-speed commands?}
As shown in Table~\ref{tab:speed_overtake}, models trained with explicit target-speed supervision consistently achieve significantly higher Speed-Adherence Scores compared to the vanilla TCP, demonstrating a basic ability to follow target-speed commands. 

\parhead{2. How do different annotation strategies affect speed adherence?}
Comparing expert demonstration and re-annotations, minimal differences in Speed-Adherence Score are observed. This suggests that \textbf{virtual target-speed provides supervision quality comparable to expert parameters}, which is particularly important for real-world datasets where expert internal parameters are unavailable. In terms of virtual speed annotation, models trained under the \textbf{virtual-short} setting achieve \textbf{marginally better} speed adherence, whereas the virtual-long setting remains comparable but slightly less stable, as shown in Fig.~\ref{fig:heatmap_speed}. This is likely because the monotonic trend-based re-annotation becomes increasingly uncertain as the extrapolation horizon extends, leading to amplified noise in the constructed target speeds.

\begin{figure}[t]
\centering
\includegraphics[width=1.0\linewidth]{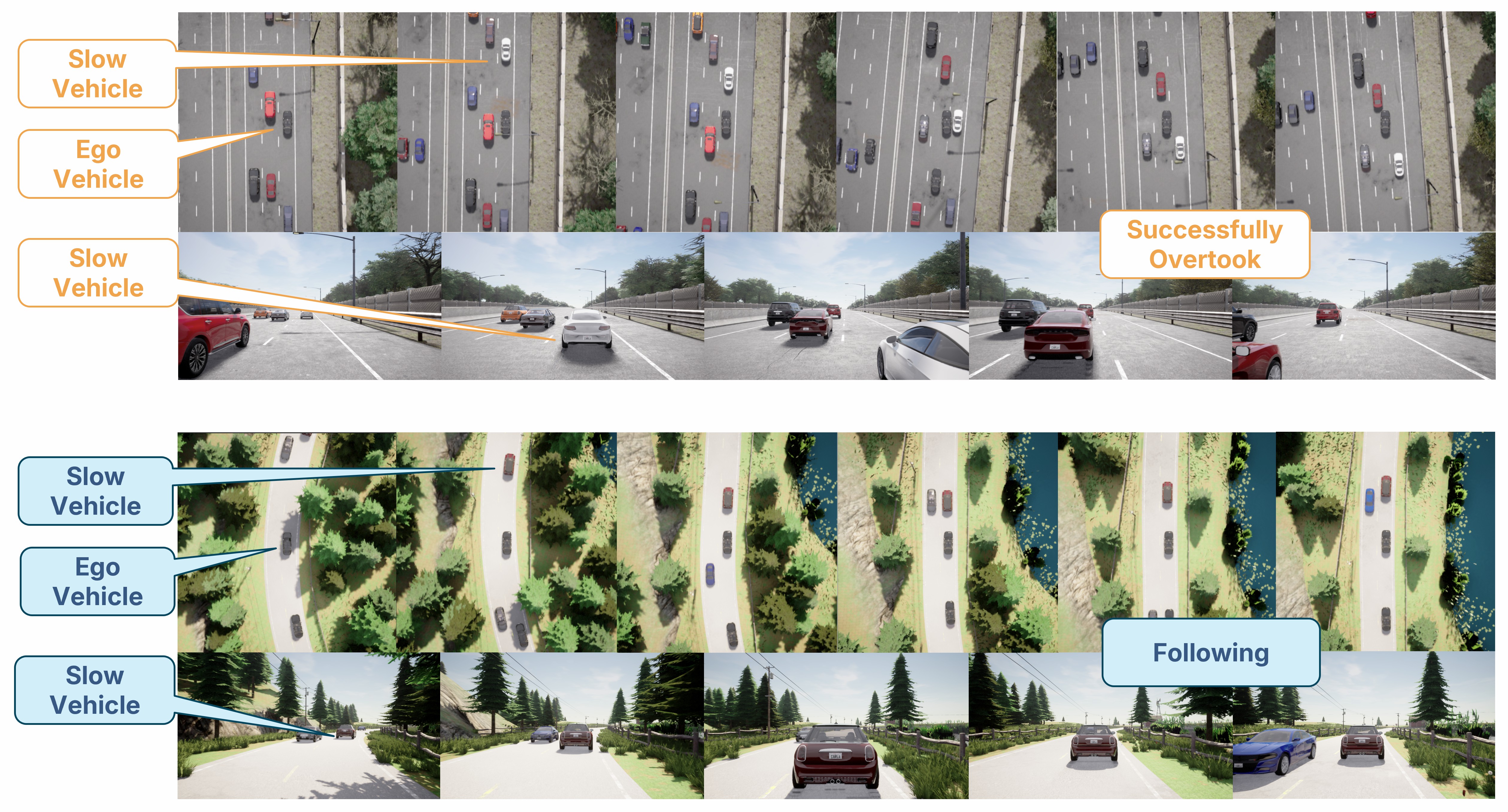}
\caption{\textbf{Visualization of \ourbaseline{} Overtaking and Following.} A successful overtaking case (upper) in which the ego vehicle passed the slow white vehicle ahead, and a successful following case (lower), in which the ego followed the vehicle till the end of the route.}
\vspace{-10pt}
\label{fig:overtake_vs_follow}
\end{figure}

\parhead{3. To what extent can the baseline follow overtaking commands?}
Models trained with \ourdataset{}-based supervision successfully exhibit distinct behaviors in response to overtake and follow commands, as evidenced by different actions under different commands and some successful overtaking attempts (Fig.~\ref{fig:overtake_vs_follow}). However, \textbf{the overtaking performance remains limited}, especially in Hard scenarios. Executing overtaking behaviors is intrinsically challenging, often requiring aggressive maneuvers that increase collision risks. In many cases, overtaking attempts lead to safety violations, reducing route completion and consequently harming the overtake score.

\begin{table}[tb!]
\centering
\small
\caption{\textbf{Closed-loop Planning Performance on 220 Route of Bench2Drive}. All models are trained under \btd{}. * denotes expert feature distillation. Non-TCP methods and distilled methods are shown in gray for reference.
\vspace{-2mm} \label{tab:b2d_closed}}
\resizebox{\linewidth}{!}{
\begin{tabular}{l|cccc}
\toprule
\textbf{Method} 
& \cellcolor{gray!15} Driving Score $\uparrow$  
& Success Rate(\%) $\uparrow$ 
& Efficiency $\uparrow$ 
& Comfortness $\uparrow$\\ 
\midrule
\g AD-MLP~\cite{ADMLP}
& \g \cellcolor{gray!15} 18.05
& \g 0.00  
& \g 48.45 
& \g 22.63  \\ 

\g UniAD-Tiny~\cite{UniAD}                               
& \g \cellcolor{gray!15} 40.73    
& \g 13.18 
& \g 123.92 
& \g 47.04  \\

\g UniAD-Base~\cite{UniAD}                            
& \g \cellcolor{gray!15} 45.81     
& \g 16.36 
& \g 129.21 
& \g 43.58  \\

\g VAD~\cite{VAD}                              
& \g \cellcolor{gray!15} 42.35     
& \g 15.00 
& \g \textbf{157.94} 
& \g 46.01  \\

\g DriveTransformer-Large \cite{DriveTransformer} 
& \g \cellcolor{gray!15} 63.46
& \g \textbf{35.01} 
& \g 100.64 
& \g 20.78  \\ 

\g ThinkTwice*~\cite{ThinkTwice}                             
& \g \cellcolor{gray!15} 62.44     
& \g 31.23  
& \g 69.33 
& \g 16.22  \\

\g DriveAdapter*~\cite{DriveAdapter}                            
& \g \cellcolor{gray!15}\textbf{64.22}     
& \g 33.08
& \g 70.22 
& \g 16.01   \\ 

\g TCP*~\cite{TCP}    
& \g \cellcolor{gray!15} 40.70     
& \g 15.00  
& \g 54.26 
& \g 47.80  \\ 

\g TCP-ctrl*                              
& \g \cellcolor{gray!15} 30.47    
& \g 7.27  
& \g 55.97 
& \g \textbf{51.51}  \\

\g TCP-traj*    
& \g \cellcolor{gray!15} 59.90
& \g 30.00
& \g 76.54 
& \g 18.08  \\

\midrule

$\text{TCP}_{\text{w/o Speed Command}}$                      
& \cellcolor{gray!15} 49.30     
& 20.45  
& 78.78 
& \textbf{22.96}  \\

\ourbaseline{} \btdshort{}

& \cellcolor{gray!15} \textbf{54.15}
& \textbf{22.73}
& 195.48 
& 20.92  \\ 

\ourbaseline{} \btdlong{}

& \cellcolor{gray!15} 51.84     
& 21.36
& \textbf{195.96}
& 22.64  \\ 

\bottomrule
\end{tabular}}
\vspace{-4mm}
\end{table}

\parhead{4. How do traditional closed-loop metrics trade off?}

\begin{itemize}[leftmargin=10pt, topsep=0pt, itemsep=1pt, partopsep=1pt, parsep=1pt,label=$\bullet$]
\item \textbf{Driving Score and Success Rate:} Speed-conditioned models show no degradation compared to the original model. On the 220 routes of Bench2Drive (Table~\ref{tab:b2d_closed}), \ourbaseline{} achieves slightly higher Driving Score and Success Rate than vanilla TCP when all of them are trained on the same \btd{} dataset without expert feature distillation, indicating that \textbf{task completion and safety remain unaffected.}

\item \textbf{Driving Smoothness (Comfort):} \ourbaseline{} shows a negligible decrease in comfort compared to the original TCP, reflecting the increased complexity of longitudinal control when adhering to target-speed commands. However, the reduction is minor and does not significantly affect driving quality.

\item \textbf{Efficiency:} Efficiency is significantly improved under speed-conditioned supervision. This improvement is mainly because speed-related tasks expose the model to multiple target speeds, which are typically higher than those learned by single-policy models without speed-command guidance.
\end{itemize}

\section{Conclusion}
\label{sec:conclusion}

We present \ourwork{}, which introduces a closed-loop benchmark for evaluating speed-conditioned autonomous driving, together with corresponding datasets and baseline implementations. By incorporating target-speed inputs and overtaking commands, as well as designing dedicated evaluation tasks, we establish a controlled setting to study speed regulation in realistic closed-loop environments.



\clearpage

%
%
\bibliographystyle{splncs04}
\bibliography{main}
\newpage

\input{appendix}

\end{document}

%% file: appendix.tex
\begin{center}
    \Large \textbf{Can Users Specify Driving Speed? \\ \ourwork: Benchmark and Baselines for Desired-Speed Conditioned Autonomous Driving} \\
    \vspace{-3mm}
\end{center}

\vspace{24px}

In the appendix, we provide additional details that were omitted from the main text. In~\autoref{sec:appendix_target_speed}, we describe the methodology for specifying target speeds for autonomous driving tasks in \ourwork{}. In~\autoref{sec:overtake_route_appendix}, we present the detailed definition and underlying mechanisms of the overtake/follow scenario implemented in our work. In~\autoref{sec:dataset_appendix}, we provide comprehensive information on the data distribution and annotation procedures of the \ourdataset{}. Finally, in~\autoref{sec:expr_appendix}, we report supplementary experimental results that complement the findings presented in the main text.

\section{Target Speed Specification}
\label{sec:appendix_target_speed}

\textbf{Formalization.} To specify the target speed in route scenarios, we extend the standard route-based scenario definition by introducing segment-wise target speed control. Each segment can be represented as a tuple:
\begin{equation}
\mathcal{S} = \{(s_i^{start}, s_i^{end}, v_i)\},
\end{equation}
where $s_i^{start}, s_i^{end} \in [0,1]$ denote normalized progress intervals along the route, and $v_i$ is the assigned target speed for that segment. Given route keypoints $\{p_k\}_{k=1}^{N}$, we compute cumulative arc-length distances:
\begin{equation}
d_k = \sum_{j=2}^{k} \|p_j - p_{j-1}\|_2,
\end{equation}
with total route length $L = d_N$. Each keypoint is thus associated with normalized progress $d_k / L$.

\parhead{Segment-to-Keypoint Speed Assignment.} For each speed segment, we convert normalized bounds to absolute arc-length:
\begin{equation}
[s_i^{start} L,\; s_i^{end} L).
\end{equation}

All keypoints whose cumulative distance falls within this interval are assigned target speed $v_i$. Unassigned keypoints inherit the most recent valid speed to ensure a fully specified speed profile along the route. This process yields a dense global speed plan:
\begin{equation}
\mathcal{P} = \{(p_k, v_k)\}_{k=1}^{N}.
\end{equation}

\parhead{Runtime Target Speed Query.} During execution, the autonomous agent retrieves the instantaneous target speed based on the ego vehicle's current location $x_{ego}$. We perform nearest-neighbor matching over $\mathcal{P}$:
\begin{equation}
v^* = v_{k^*}, \quad
k^* = \arg\min_k \|x_{ego} - p_k\|_2.
\end{equation}

The selected $v^*$ serves as the planned target speed at runtime.

\begin{figure}[t]
    \centering
    \includegraphics[width=1.0\linewidth]{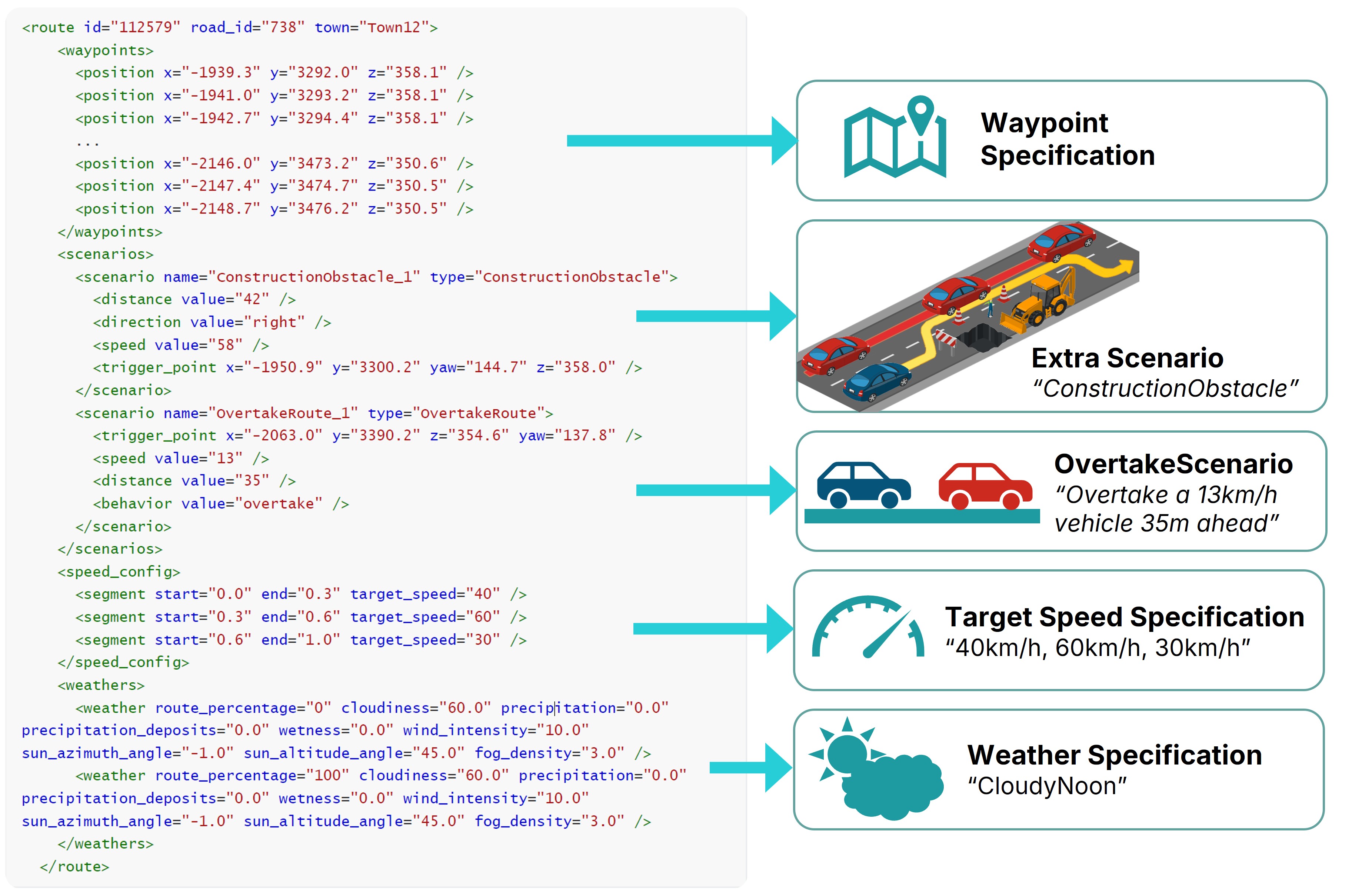}
    \caption{\textbf{Example of a configuration XML file.} The configuration specifies route waypoints, traffic scenarios, target speed settings, and weather conditions, \etc.}
    \label{fig:xml_example}
\end{figure}

\section{Overtake/Follow Scenario Implementation}
\label{sec:overtake_route_appendix}

To evaluate whether an agent complies with efficiency-related lateral instructions (\texttt{overtake} or \texttt{follow}) under controlled traffic conditions, we implement a custom scenario, OvertakeRoute, within the CARLA ScenarioRunner framework.

\parhead{Scenario Overview.} The scenario introduces a slow-moving vehicle ahead of the ego vehicle along a predefined route. Depending on the given command:

\begin{itemize}[leftmargin=10pt, topsep=0pt, itemsep=1pt, partopsep=1pt, parsep=1pt,label=$\bullet$]
    \item Under the \texttt{overtake} command, the ego vehicle must pass the front vehicle and establish a lead.
    \item Under the \texttt{follow} command, the ego vehicle must remain behind the front vehicle, even if it is moving at a low speed.
\end{itemize}


\parhead{Configuration Parameters.} We extend the standard scenario configuration files with additional parameters:

\begin{itemize}[leftmargin=10pt, topsep=0pt, itemsep=1pt, partopsep=1pt, parsep=1pt,label=$\bullet$]
    \item \texttt{speed}: target speed of the front vehicle,
    \item \texttt{distance}: initial longitudinal distance between ego and front vehicle,
    \item \texttt{behavior}: command type (\texttt{overtake} or \texttt{follow}),
    \item \texttt{frequency}: spawning frequency of oncoming vehicles (two-way variant only).
\end{itemize}

These parameters are parsed at initialization and allow flexible difficulty adjustment across different route segments. An example of configuration file is provided in Fig.~\ref{fig:xml_example}.

\parhead{Actor Initialization.} At runtime, a front vehicle is spawned at a waypoint located a configurable distance ahead of the ego vehicle's trigger point. The vehicle follows the route at the specified cruising speed. \textbf{No other vehicle is spawned in the same lane to prevent violation of this scenario.}


\parhead{Scenario Behavior.} (1) Trigger Condition, the scenario activates when the ego vehicle enters a predefined trigger region near the route waypoint. (2) Front Vehicle Behavior, the front vehicle drives forward toward a distant waypoint at the configured cruising speed and continues indefinitely unless termination conditions are met.

\parhead{Termination Conditions.} The scenario terminates when one of the following occurs: (1) the ego vehicle successfully moves ahead of the front vehicle (overtake completed), (2) a predefined timeout is reached, (3) the route evaluation ends.

\section{Dataset Details}
\label{sec:dataset_appendix}

\subsection{Data Distribution}

\begin{figure}[t]
    \centering
    \begin{subfigure}[b]{0.44\linewidth}
        \centering
        \includegraphics[width=\linewidth]{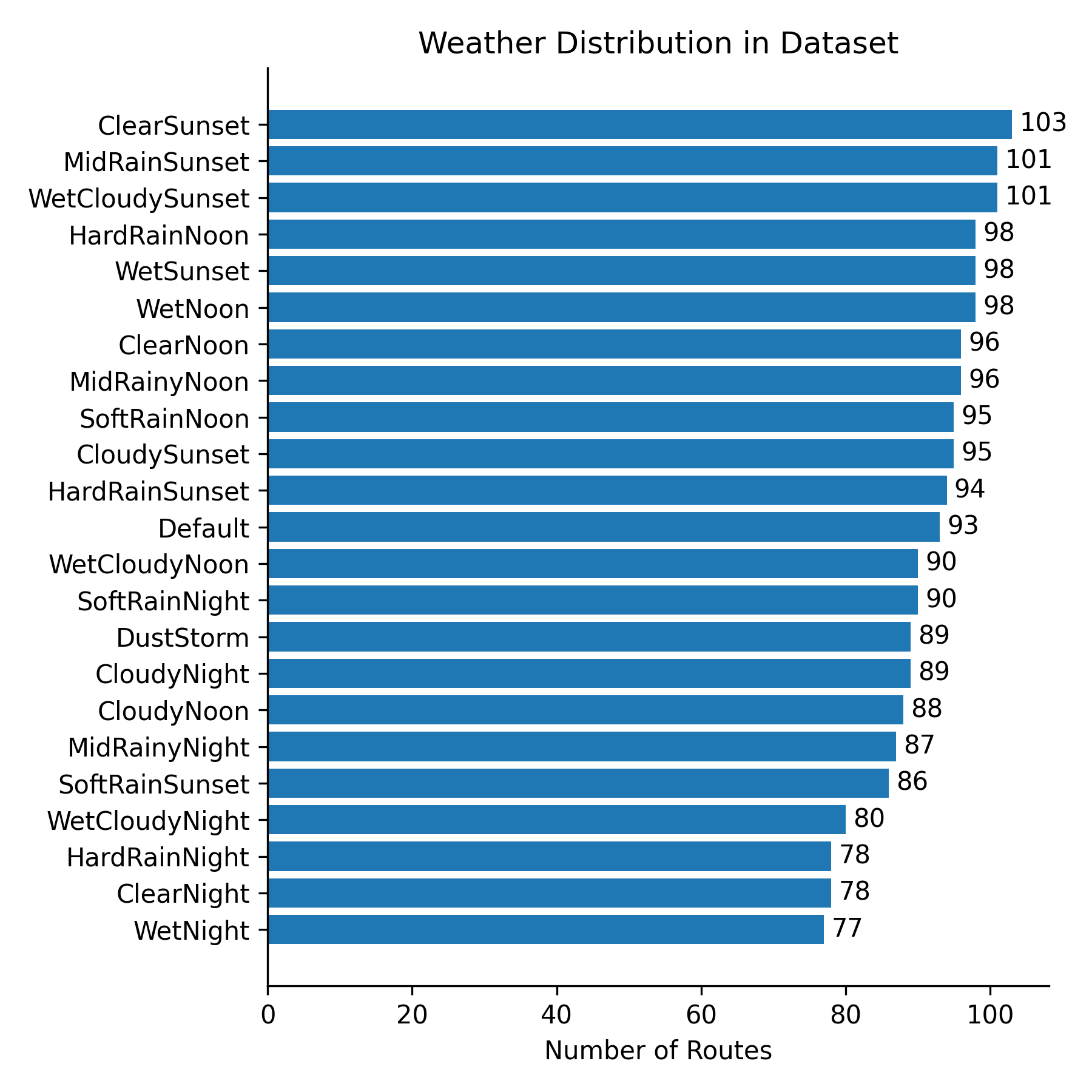}
        \caption{Weather distribution in \ourdataset{}.}
        \label{fig:weather_dist}
    \end{subfigure}
    \hspace{0.1\linewidth}
    \begin{subfigure}[b]{0.44\linewidth}
        \centering
        \includegraphics[width=\linewidth]{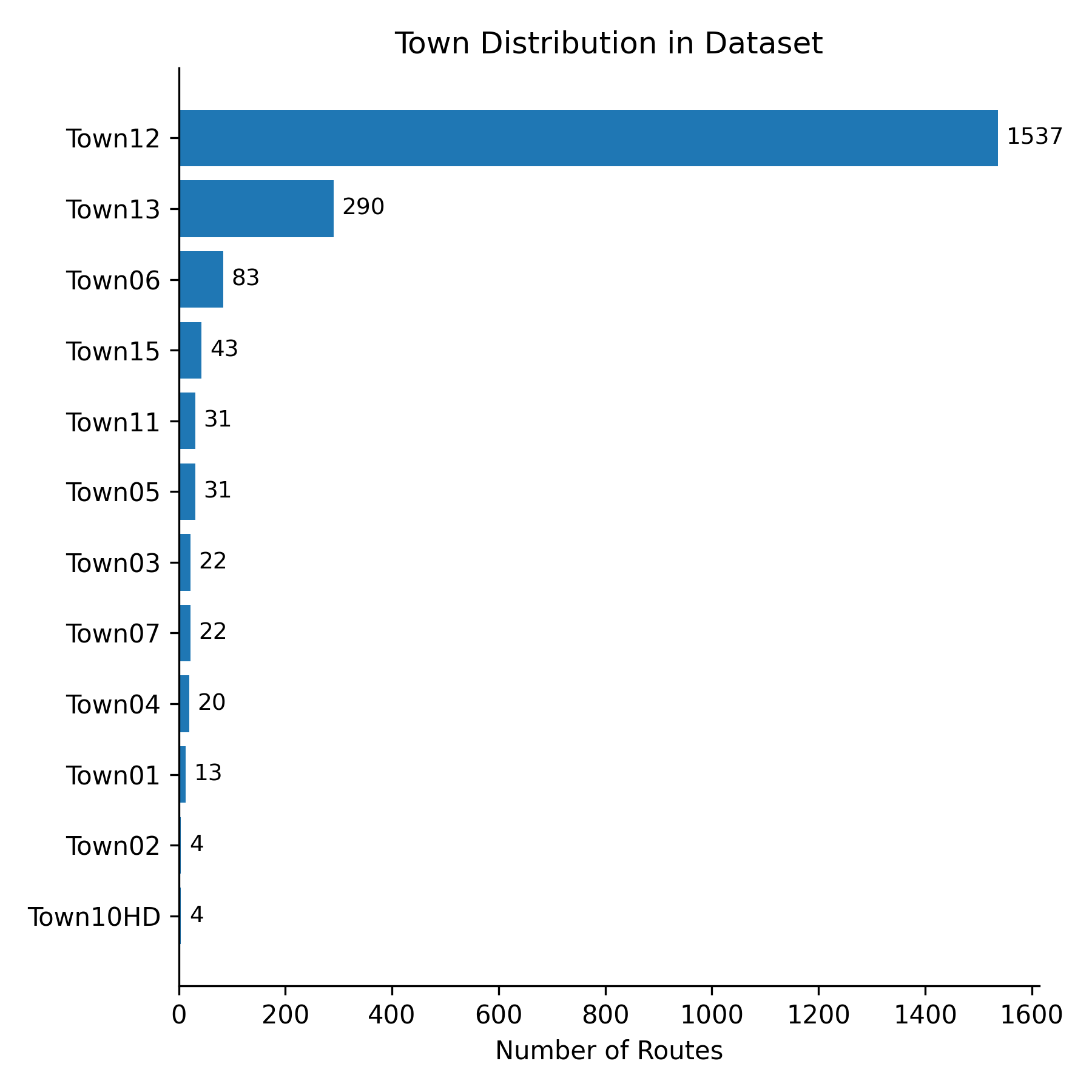}
        \caption{Town distribution in \ourdataset{}.}
        \label{fig:town_dist}
    \end{subfigure}
    \caption{(a) \textbf{Distribution of weather conditions in \ourdataset{}.} It covers all predefined weathers in the simulator, uniformly distributed. (b) \textbf{Distribution of towns in \ourdataset{}.} Route counts roughly reflect each town’s size and number of routes.}
    \label{fig:weather_town_dist}
\end{figure}

\begin{figure}[t]
    \centering
    \begin{subfigure}[t]{0.52\linewidth}
        \centering
        \includegraphics[width=\linewidth]{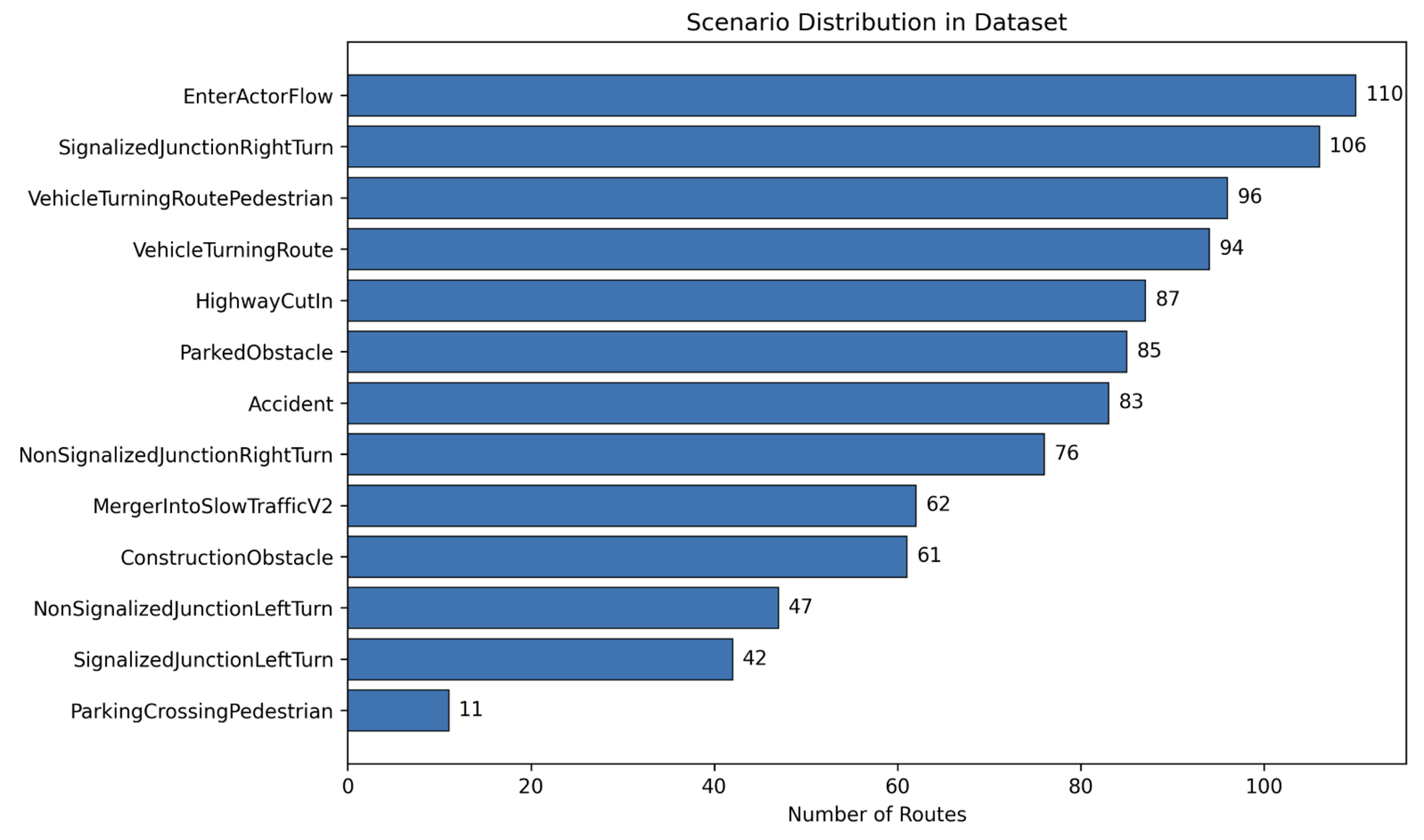}
        \caption{Scenario-wise distribution of hard routes.}
        \label{fig:scenario_dist}
    \end{subfigure}
    \hspace{0.04\linewidth}
    \begin{subfigure}[t]{0.42\linewidth}
        \centering
        \includegraphics[width=\linewidth]{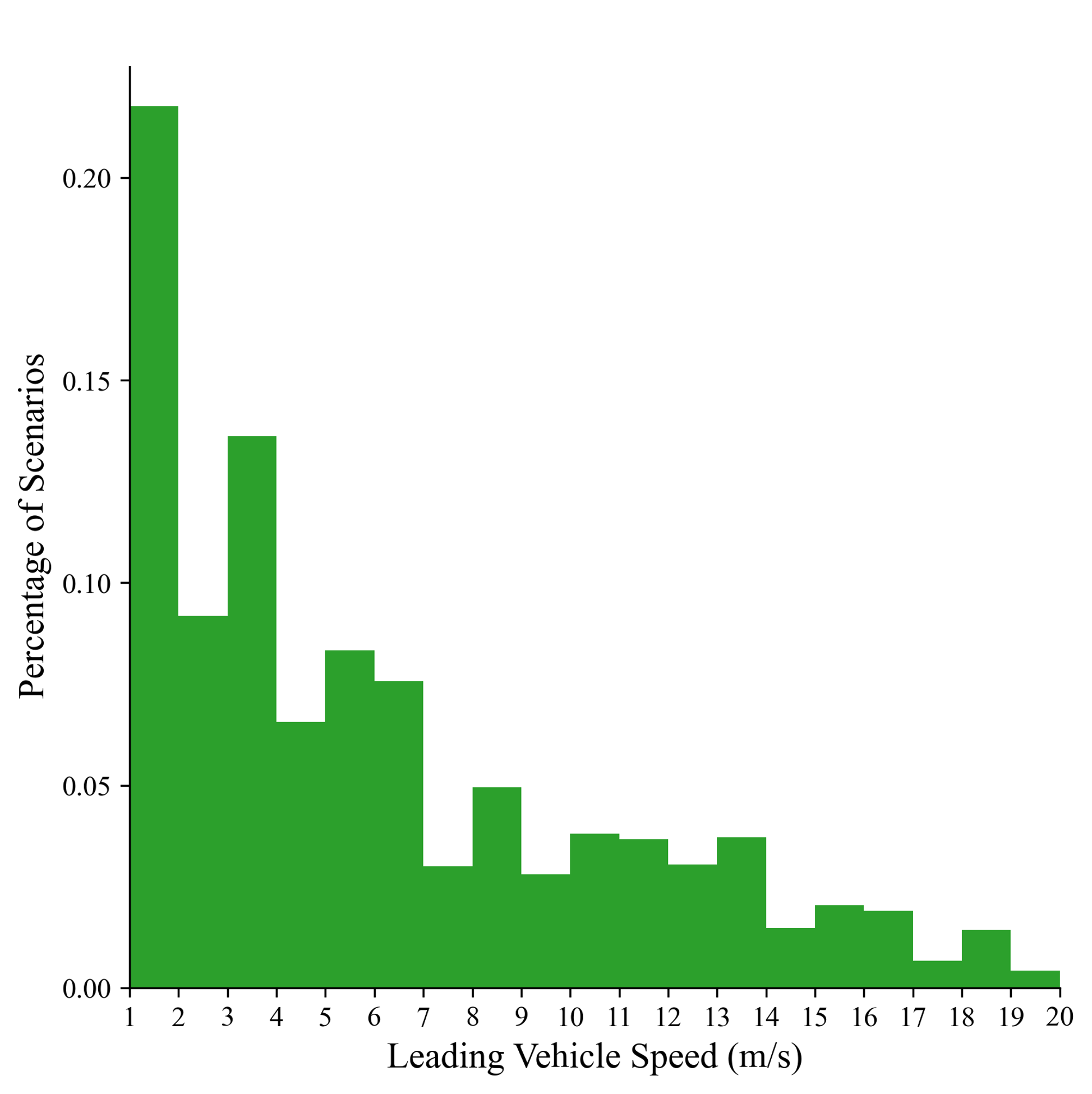}
        \caption{Leading vehicle speed distribution in overtake/follow scenarios.}
        \label{fig:overtake_speed_distribution}
    \end{subfigure}
    \caption{(a) \textbf{Scenario-wise distribution of hard routes in \ourdataset{}.} Covering the most challenging scenarios in the simulator.
    (b) \textbf{Distribution of leading vehicle speeds in overtake/follow scenarios within \ourdataset{}.} This speed is sampled from a small range below the ego vehicle speed, resulting in a distribution biased toward lower speeds.}
    \label{fig:combined_distribution}
\end{figure}

In the main section, we have demonstrated that the target speed distribution in \ourdataset{} is balanced. Furthermore, the front vehicle speed $v_f$ is subsequently sampled from a uniform distribution over $[\epsilon, v_t - \delta]$, where $\epsilon$ denotes a small lower-bound threshold and $\delta$ represents a small speed margin ensuring that the front vehicle remains slower than the ego vehicle. Consequently, the marginal distribution of $v_f$ is biased toward lower speeds, as smaller speeds can be sampled for a wider range of ego target speeds, as shown in Fig.~\ref{fig:overtake_speed_distribution}.
In addition to these speed-related variables, the weathers in \ourdataset{} are also uniformly distributed, as shown in Fig.~\ref{fig:weather_dist}. 

The distribution of towns is shown in Fig.~\ref{fig:town_dist} - route counts roughly correlate with the area and number of routes in each town, so most routes are located in Town12.  
In hard routes, a scenario from CARLA Leaderboard v2 occurs, forcing the autonomous driving agent to handle complex corner cases while adhering to style-related commands. The distribution of these scenarios is shown in Fig.~\ref{fig:scenario_dist}.

\subsection{Virtual Target Speed Annotation}
\label{sec:virtual_speed}



In addition to the target speed from the expert's internal parameter, which is impossible to derive from real-life datasets, \ourdataset{} introduces a \textbf{virtual target speed} for each frame to facilitate style-aware driving tasks.

\parhead{Tendency Speed.}  
We first define the \textbf{tendency speed} $v^{\text{tend}}_t$ at frame $t$ as the maximal (or minimal) speed in a short future horizon that maintains the current monotonic trend:

\begin{equation}
v^{\text{tend}}_t =
\begin{cases}
\max\limits_{i \in [1, F]} v_{t+i}, & \text{if } v_{t+1} > v_t \quad (\text{acceleration trend}) \\
\min\limits_{i \in [1, F]} v_{t+i}, & \text{if } v_{t+1} < v_t \quad (\text{deceleration trend}) \\
v_t, & \text{otherwise}
\end{cases}
\end{equation}

where $v_{t+i}$ is the ego speed at frame $t+i$, and $F$ is the number of future frames considered (in our implementation, $F = 40$).

\parhead{Virtual Target Speed.}  
Given the current tendency speed $v^{\text{tend}}_t$ and the previous tendency speed $v^{\text{tend}}_{t-1}$, the virtual target speed $v^{\text{virt}}_t$ is computed by linearly extrapolating the trend over a randomized short horizon $\Delta t$:

\begin{equation}
v^{\text{virt}}_t = \max \Big( v^{\text{tend}}_t + \Delta v_t, 0 \Big), \quad
\Delta v_t = (v^{\text{tend}}_t - v^{\text{tend}}_{t-1}) \cdot \text{FPS} \cdot r, \quad r \sim \mathcal{U}(T_{\min}, T_{\max})
\end{equation}

Here, $\text{FPS}$ is the frame rate of the dataset, which is $10$; $r$ is a random factor uniformly sampled in $[T_{\min}, T_{\max}]$, and the resulting $\Delta v_t$ is clipped to lie within $$[-\text{MAX\_EXTEND}, \text{MAX\_EXTEND}]$$ to prevent unrealistic speed jumps.

\parhead{Short vs Long Virtual Target Speed.} To support different experimental setups, \ourdataset{} provides two versions of virtual target speed annotations:

\begin{itemize}[leftmargin=10pt, topsep=0pt, itemsep=1pt, partopsep=1pt, parsep=1pt,label=$\bullet$]
    \item \textbf{Long}: uses $\text{MAX\_EXTEND} = 10.0 \,\mathrm{m/s}$ and $T_{\max} = 3.0\,\mathrm{s}$, which emphasizes long-horizon extrapolation and captures extended acceleration/deceleration trends.
    \item \textbf{Short}: uses $\text{MAX\_EXTEND} = 3.0 \,\mathrm{m/s}$ and $T_{\max} = 1.5\,\mathrm{s}$, which focuses on short-term variations and preserves finer temporal detail.
\end{itemize}

\section{Experiment Details}
\label{sec:expr_appendix}

\parhead{Training.} All baseline models are trained on a single NVIDIA A100 GPU using PyTorch Lightning. The optimizer is Adam with an initial learning rate of ($1\times10^{-4}$) and weight decay ($1\times10^{-7}$). The learning rate is scheduled using a StepLR scheduler with a decay factor of 0.5 every 30 epochs. All models are trained for 60 epochs with a batch size of 300.

\begin{table*}[t]
\centering
\setlength{\tabcolsep}{5pt}
\caption{\textbf{Speed-Adherence Score and Overtake Score} on 48 evaluation routes of \ourwork{}. Metrics are reported for All(A), Easy (E), Medium (M), and Hard (H). The best result of each score is highlighted in bold.}
\label{tab:speed_overtake_full}
\resizebox{1.0\textwidth}{!}{
\begin{tabular}{ll|cccc|cccc}
\toprule
\multirow{2}{*}{\textbf{Model}} 
& \multirow{2}{*}{\textbf{Dataset}}
& \multicolumn{4}{c|}{\textbf{Speed Adherence Score $\uparrow$}}
& \multicolumn{4}{c}{\textbf{Overtake Score $\uparrow$}} \\
\cmidrule(lr){3-6}
\cmidrule(lr){7-10}
& 
& \cellcolor{gray!15}\textbf{A} & E & M & H
& \cellcolor{gray!15}\textbf{A} & E & M & H \\

\midrule

$\text{TCP}_{\text{w/o Speed Command}}$ & \btd{}
& \cellcolor{gray!15}41.36 & 42.03 & 40.72 & 41.33
& \cellcolor{gray!15}37.50 & - & 50.00 & \textbf{25.00} \\

TCP-Speed & \btdshort{}
& \cellcolor{gray!15}67.95 & 71.01 & 69.63 & 63.21
& \cellcolor{gray!15}31.25 & - & 50.00 & 12.50 \\

TCP-Speed & \btdlong{}
& \cellcolor{gray!15}65.61 & 67.80 & 65.69 & 63.34
& \cellcolor{gray!15}31.25 & - & 43.75 & 18.75 \\

\midrule

$\text{TCP}_{\text{w/o Speed Command}}$ & \btd{}+\ourdataset{}
& \cellcolor{gray!15}41.60 & 42.11 & 40.98 & 41.70
& \cellcolor{gray!15}25.00 & - & 50.00 & 0.00 \\

TCP-Speed & \btdshort{}+\virtualshort{}
& \cellcolor{gray!15}\textbf{71.65} & 77.33 & \textbf{72.63} & 64.99
& \cellcolor{gray!15}34.38 & - & 50.00 & 18.75 \\

TCP-Speed & \btdlong{}+\virtuallong{}
& \cellcolor{gray!15}71.21 & \textbf{79.02} & 69.16 & \textbf{66.40}
& \cellcolor{gray!15}34.38 & - & 50.00 & 18.75 \\

\midrule

$\text{TCP}_{\text{w/o Speed Command}}$ & \ourdataset{}
& \cellcolor{gray!15}41.54 & 42.00 & 40.95 & 41.67
& \cellcolor{gray!15}18.75 & - & 37.50 & 0.00 \\

TCP-Speed & \virtualshort{}
& \cellcolor{gray!15}69.23 & 76.18 & 68.71 & 62.81
& \cellcolor{gray!15}\textbf{40.63} & - & \textbf{56.25} & \textbf{25.00} \\

TCP-Speed & \virtuallong{}
& \cellcolor{gray!15}68.36 & 73.17 & 67.93 & 63.98
& \cellcolor{gray!15}\textbf{40.63} & - & \textbf{56.25} & \textbf{25.00} \\

TCP-Speed & \expertdata{}
& \cellcolor{gray!15}68.79 & 76.80 & 65.38 & 64.20
& \cellcolor{gray!15}21.88 & - & 37.50 & 6.25 \\

\bottomrule
\end{tabular}
}
\end{table*}

\begin{table*}[t]
\centering
\setlength{\tabcolsep}{4pt}
\caption{\textbf{Overall closed-loop performance on traditional metrics (Driving Score, Success Rate, Efficiency, Comfortness) on 48 evaluation routes of \ourwork{}.} 
The best overall result of each score is highlighted in bold.}
\label{tab:overall_traditional_metrics}
\resizebox{\textwidth}{!}{
\begin{tabular}{llcccc}
\toprule
\textbf{Model} & \textbf{Dataset} 
& \makecell{\textbf{Driving Score} $\uparrow$} 
& \makecell{\textbf{Success Rate (\%)} $\uparrow$} 
& \textbf{Efficiency $\uparrow$} 
& \makecell{\textbf{Comfortness} $\uparrow$} \\

\midrule

$\text{TCP}_{\text{w/o Speed Command}}$ & \btd{} 
& \textbf{56.52} & 20.83 & 44.39 & 24.79 \\

TCP-Speed & \btdshort{}
& 40.23 & 20.83 & 158.32 & 30.27 \\

TCP-Speed & \btdlong{}
& 45.18 & 27.08 & 145.22 & 29.67 \\

\midrule

$\text{TCP}_{\text{w/o Speed Command}}$ & \btd{}+\ourdataset{} 
& 52.71 & 27.08 & 52.17 & 33.00 \\

TCP-Speed & \btdshort{}+\virtualshort{}
& 30.19 & 16.67 & 158.32 & 30.27 \\

TCP-Speed & \btdlong{}+\virtuallong{}
& 48.40 & \textbf{31.25} & 145.22 & 29.67 \\

\midrule

$\text{TCP}_{\text{w/o Speed Command}}$ & \ourdataset{} 
& 45.13 & 18.75 & 48.67 & 36.34 \\

TCP-Speed & \virtualshort{}
& 29.66 & 18.75 & 169.84 & 35.57 \\

TCP-Speed & \virtuallong{}
& 28.37 & 14.58 & \textbf{177.21} & 38.37 \\

TCP-Speed & \expertdata{}
& 34.84 & 25.00 & 139.37 & \textbf{39.83} \\
\bottomrule
\end{tabular}
}
\end{table*}

\begin{table}[tb!]
\centering
\small
\caption{\textbf{Closed-loop Planning Performance on 220 routes of Bench2Drive}. * denotes expert feature distillation. Non-TCP methods and expert-feature-distilled methods are shown in gray for reference.
\vspace{-2mm} \label{tab:220_results}}
\resizebox{\linewidth}{!}{
\begin{tabular}{l|cccc}
\toprule
\textbf{Method} 
& \cellcolor{gray!15} Driving Score $\uparrow$  
& Success Rate(\%) $\uparrow$ 
& Efficiency $\uparrow$ 
& Comfortness $\uparrow$\\ 
\midrule
\g AD-MLP
& \g \cellcolor{gray!15} 18.05
& \g 0.00  
& \g 48.45 
& \g 22.63  \\ 

\g UniAD-Tiny                              
& \g \cellcolor{gray!15} 40.73    
& \g 13.18 
& \g 123.92 
& \g 47.04  \\

\g UniAD-Base                            
& \g \cellcolor{gray!15} 45.81     
& \g 16.36 
& \g 129.21 
& \g 43.58  \\

\g VAD                             
& \g \cellcolor{gray!15} 42.35     
& \g 15.00 
& \g \textbf{157.94} 
& \g 46.01  \\

\g DriveTransformer-Large
& \g \cellcolor{gray!15} 63.46
& \g \textbf{35.01} 
& \g 100.64 
& \g 20.78  \\ 

\g ThinkTwice*                           
& \g \cellcolor{gray!15} 62.44     
& \g 31.23  
& \g 69.33 
& \g 16.22  \\

\g DriveAdapter*                         
& \g \cellcolor{gray!15}\textbf{64.22}     
& \g 33.08
& \g 70.22 
& \g 16.01   \\ 

\g TCP*  
& \g \cellcolor{gray!15} 40.70     
& \g 15.00  
& \g 54.26 
& \g 47.80  \\ 

\g TCP-ctrl*                              
& \g \cellcolor{gray!15} 30.47    
& \g 7.27  
& \g 55.97 
& \g \textbf{51.51}  \\

\g TCP-traj*    
& \g \cellcolor{gray!15} 59.90
& \g 30.00
& \g 76.54 
& \g 18.08  \\

\midrule

$\text{TCP}_{\text{w/o Speed Command}}$                      
& \cellcolor{gray!15} 49.30     
& 20.45  
& 78.78 
& \textbf{22.96}  \\

\ourbaseline{} \btdshort{}

& \cellcolor{gray!15} \textbf{54.15}
& \textbf{22.73}
& 195.48 
& 20.92  \\ 

\ourbaseline{} \btdlong{}

& \cellcolor{gray!15} 51.84     
& 21.36
& \textbf{195.96}
& 22.64  \\ 

\midrule

$\text{TCP}_{\text{w/o Speed Command}}$   \btd{}+\ourdataset{}
& \cellcolor{gray!15} \textbf{55.23}     
& \textbf{25.45}
& 77.69 
& \textbf{27.26}  \\

\ourbaseline{} \btdshort{}+\virtualshort{}

& \cellcolor{gray!15} 53.66     
& 22.73
& 192.17 
& 21.59  \\ 

\ourbaseline{} \btdlong{}+\virtuallong{}

& \cellcolor{gray!15} 54.80
& 24.09
& \textbf{195.63} 
& 24.14  \\ 

\bottomrule
\end{tabular}}
\end{table}

\begin{table}[tb!]
\centering
\small
\caption{\textbf{Multi-Ability Results of E2E-AD Methods on Bench2Drive220 routes.} * denotes expert feature distillation. Non-TCP methods and expert-feature-distilled methods are shown in gray for reference.\vspace{-2mm}\label{tab:multi_ability_full}}
\resizebox{\linewidth}{!}{
\begin{tabular}{l|ccccc|c}
\toprule
\textbf{Method} & \textbf{Merging} & \textbf{Overtaking} & \textbf{Emergency Brake} & \textbf{Give Way} & \textbf{Traffic Sign} & \textbf{Mean} \\ 
\midrule
\g AD-MLP        & \g 0.00  & \g 0.00   & \g 0.00   & \g 0.00   & \g 0.00   & \g \cellcolor{gray!15}0.00  \\
\g UniAD-Tiny   & \g 7.04  & \g 10.00  & \g 21.82  & \g 20.00  & \g 14.61  & \g \cellcolor{gray!15}14.69 \\
\g UniAD-Base   & \g 12.16 & \g 20.00  & \g 23.64  & \g 10.00  & \g 13.89  & \g \cellcolor{gray!15}15.94 \\
\g VAD            & \g 7.14  & \g 20.00  & \g 16.36  & \g 20.00  & \g 20.22  & \g \cellcolor{gray!15}16.75 \\
\g DriveTransformer-Large & \g \textbf{17.57} & \g \textbf{35.00} & \g 48.36 & \g 40.00 & \g \textbf{52.10} & \g \cellcolor{gray!15}\textbf{38.60} \\ 
\g DriveAdapter* & \g 14.55 & \g 22.61 & \g \textbf{54.04} & \g \textbf{50.00} & \g 50.45 & \g \cellcolor{gray!15}38.33 \\
\g ThinkTwice*   & \g 13.72 & \g 22.93 & \g 52.99 & \g \textbf{50.00} & \g 47.78 & \g \cellcolor{gray!15}37.48 \\
\g TCP*     & \g 16.18 & \g 20.00 & \g 20.00 & \g 10.00 & \g 6.99  & \cellcolor{gray!15}\g 14.63 \\
\g TCP-ctrl*                              & \g 10.29 & \g 4.44  & \g 10.00 & \g 10.00 & \g 6.45  & \cellcolor{gray!15}\g 8.23  \\
\g TCP-traj*                              & \g 12.50 & \g 24.29 & \g 51.67 & \g 40.00 & \g 46.28 & \cellcolor{gray!15}\g 34.22 \\
\midrule
$\text{TCP}_{\text{w/o Speed Command}}$ & 17.14 & 6.67  & \textbf{40.00} & \textbf{50.00} & 28.72 & \cellcolor{gray!15}28.51 \\
\ourbaseline{} \btdshort{}             & \textbf{22.50} & 6.67  & 35.00 & \textbf{50.00} & 41.05 & \cellcolor{gray!15}\textbf{31.04} \\ 
\ourbaseline{} \btdlong{}              & 21.25 & \textbf{8.89} & 35.00 & 40.00 & \textbf{45.79} & \cellcolor{gray!15}30.19 \\ 
\midrule
$\text{TCP}_{\text{w/o Speed Command}}$ \btd{}+\ourdataset{} & 19.30 & 11.11 & \textbf{45.00} & 40.00 & \textbf{43.42} & \cellcolor{gray!15}\textbf{31.77} \\
\ourbaseline{} \btdshort{}+\virtualshort{} & 18.75 & \textbf{15.56} & 35.00 & \textbf{50.00} & 38.95 & \cellcolor{gray!15}31.65 \\
\ourbaseline{} \btdlong{}+\virtuallong{} & \textbf{23.75} & 8.89 & 35.00 & 40.00 & 38.94 & \cellcolor{gray!15}29.32 \\
\bottomrule
\end{tabular}}
\vspace{-4mm}
\end{table}

\parhead{Speed-Related Task Evaluations.} We evaluate the baseline models on the 48 routes of \ourwork{}. In Table~\ref{tab:speed_overtake_full}, we report the performance of the 48 routes in \ourwork{}, broken down by difficulty level across the primary evaluation metrics. The results indicate that

\begin{itemize}[leftmargin=10pt, topsep=0pt, itemsep=1pt, partopsep=1pt, parsep=1pt,label=$\bullet$]
\item Compared to the vanilla TCP, our baseline achieves higher Speed and Overtake Scores but slightly lower Driving Score and Success Rate. This drop is attributed to the increased complexity introduced by diverse target-speed and overtaking commands, which pose additional challenges for target-speed-aware models.
\item Models trained on the re-annotated datasets demonstrate comparable speed adherence as models trained on expert demonstrations.
\item Models trained on the re-annotated datasets achieve even better overtake scores and efficiency than those trained on expert demonstrations. This suggests that the model may struggle to capture the implicit relationship between the expert-annotated target speed and the ego vehicle speed in complex scenarios such as overtaking. As a result, the learned policy tends to converge to an averaged behavior across scenarios sharing the same target speed, leading to a lower speed tendency.
\item Models trained on expert demonstrations achieve slightly higher driving scores and success rates. Expert demonstrations reflect more complex, context-dependent speed choices, resulting in less direct adherence to target speeds but better overall driving performance, while re-annotated speeds are derived straightforwardly from observed trajectories and may overfit.
\end{itemize}

\parhead{Traditional AD Metrics Evaluations.} Since evaluating the full set of 220 routes in the original Bench2Drive benchmark requires a substantial amount of time, only a subset of baseline methods were evaluated on the complete set. Table~\ref{tab:220_results} reports the results on four core metrics—Driving Score, Success Rate, Efficiency, and Comfort—while Table~\ref{tab:multi_ability_full} presents the multi-ability evaluation results. From the Bench2Drive-220 evaluation, we observe that in both dataset configurations, including \btd{}-only and \btd{} + \ourdataset{}, 

\begin{itemize}[leftmargin=10pt, topsep=0pt, itemsep=1pt, partopsep=1pt, parsep=1pt,label=$\bullet$]
\item \ourbaseline{} achieves Driving Score and Success Rate comparable to the original TCP, indicating that speed conditioning preserves the core safety-related capabilities. 
\item Style-aware models consistently obtain higher Efficiency scores, suggesting reduced overly conservative behaviors compared to the single-style policy implicitly on traditional datasets.
\item The Comfort metric remains largely comparable to the original model, indicating that following target speeds only slightly affects driving smoothness.
\item The multi-ability evaluation shows competitive performance on challenging maneuvers such as merging and overtaking.
\end{itemize}

%% file: main.bbl
\begin{thebibliography}{10}
\providecommand{\url}[1]{\texttt{#1}}
\providecommand{\urlprefix}{URL }
\providecommand{\doi}[1]{https://doi.org/#1}

\bibitem{aledhari2023motion}
Aledhari, M., Rahouti, M., Qadir, J., Qolomany, B., Guizani, M., Al-Fuqaha, A.: Motion comfort optimization for autonomous vehicles: Concepts, methods, and techniques. IEEE Internet of Things Journal  \textbf{11}(1),  378--402 (2024). \doi{10.1109/JIOT.2023.3287489}

\bibitem{beisswenger2024pdmlite}
Beißwenger, J.: {PDM-Lite}: A rule-based planner for carla leaderboard 2.0. \url{https://github.com/OpenDriveLab/DriveLM/tree/DriveLM-CARLA} (2024)

\bibitem{NuScenes}
Caesar, H., Bankiti, V., Lang, A.H., Vora, S., Liong, V.E., Xu, Q., Krishnan, A., Pan, Y., Baldan, G., Beijbom, O.: { NuScenes: A Multimodal Dataset for Autonomous Driving}. In: 2020 IEEE/CVF Conference on Computer Vision and Pattern Recognition (CVPR). pp. 11618--11628. IEEE Computer Society, Los Alamitos, CA, USA (Jun 2020). \doi{10.1109/CVPR42600.2020.01164}, \url{https://doi.ieeecomputersociety.org/10.1109/CVPR42600.2020.01164}

\bibitem{cao2025adapt}
Cao, Y., Jiang, Y., Zeng, X.: Adaptive game-theoretic decision-making with driving style recognition for autonomous vehicles in uninterrupted traffic flows at intersections. Robotics and Autonomous Systems  \textbf{194},  105180 (2025). \doi{https://doi.org/10.1016/j.robot.2025.105180}, \url{https://www.sciencedirect.com/science/article/pii/S0921889025002775}

\bibitem{chen2024metafollower}
Chen, X., Chen, K., Zhu, M., Yang, H.F., Shen, S., Wang, X., Wang, Y.: Metafollower: Adaptable personalized autonomous car following. Transportation Research Part C: Emerging Technologies  \textbf{169},  104872 (2024). \doi{https://doi.org/10.1016/j.trc.2024.104872}, \url{https://www.sciencedirect.com/science/article/pii/S0968090X24003930}

\bibitem{codevilla2018end}
Codevilla, F., Müller, M., López, A., Koltun, V., Dosovitskiy, A.: End-to-end driving via conditional imitation learning. In: 2018 IEEE International Conference on Robotics and Automation (ICRA). pp. 4693--4700 (2018). \doi{10.1109/ICRA.2018.8460487}

\bibitem{cui2024personalized}
Cui, C., Yang, Z., Zhou, Y., Ma, Y., Lu, J., Li, L., Chen, Y., Panchal, J., Wang, Z.: Personalized autonomous driving with large language models: Field experiments. In: 2024 IEEE 27th International Conference on Intelligent Transportation Systems (ITSC). pp. 20--27 (2024). \doi{10.1109/ITSC58415.2024.10919978}

\bibitem{cui2024onboard}
Cui, C., Yang, Z., Zhou, Y., Peng, J., Park, S.Y., Zhang, C., Ma, Y., Cao, X., Ye, W., Feng, Y., Panchal, J., Li, L., Chen, Y., Wang, Z.: On-board vision-language models for personalized autonomous vehicle motion control: System design and real-world validation (2024), \url{https://arxiv.org/abs/2411.11913}

\bibitem{dauner23aPDM-Closed}
Dauner, D., Hallgarten, M., Geiger, A., Chitta, K.: Parting with misconceptions about learning-based vehicle motion planning. In: Proceedings of The 7th Conference on Robot Learning. Proceedings of Machine Learning Research, vol.~229, pp. 1268--1281. PMLR (06--09 Nov 2023)

\bibitem{NAVSIM}
Dauner, D., Hallgarten, M., Li, T., Weng, X., Huang, Z., Yang, Z., Li, H., Gilitschenski, I., Ivanovic, B., Pavone, M., Geiger, A., Chitta, K.: Navsim: Data-driven non-reactive autonomous vehicle simulation and benchmarking. In: Advances in Neural Information Processing Systems. vol.~37, pp. 28706--28719. Curran Associates, Inc. (2024). \doi{10.52202/079017-0902}

\bibitem{CARLA}
Dosovitskiy, A., Ros, G., Codevilla, F., Lopez, A., Koltun, V.: {CARLA}: {An} open urban driving simulator. In: Proceedings of the 1st Annual Conference on Robot Learning. Proceedings of Machine Learning Research, vol.~78, pp. 1--16. PMLR (13--15 Nov 2017)

\bibitem{ORION}
Fu, H., Zhang, D., Zhao, Z., Cui, J., Liang, D., Zhang, C., Zhang, D., Xie, H., Wang, B., Bai, X.: Orion: A holistic end-to-end autonomous driving framework by vision-language instructed action generation. In: Proceedings of the IEEE/CVF International Conference on Computer Vision (ICCV). pp. 24823--24834 (October 2025)

\bibitem{carla_garage_common_mistakes}
Group, A.V.: Common mistakes in benchmarking ad. \url{https://github.com/autonomousvision/carla_garage/blob/leaderboard_2/docs/common_mistakes_in_benchmarking_ad.md} (2023), accessed: 2025-04-25

\bibitem{han2024words}
Han, X., Chen, X., Cai, Z., Cai, P., Zhu, M., Chu, X.: From words to wheels: Automated style-customized policy generation for autonomous driving (2024), \url{https://arxiv.org/abs/2409.11694}

\bibitem{han2024vehicle}
Hao, J., Xie, H., Guo, F., Chen, Y., Song, K.: Vehicle trajectory prediction with driving style identification and intention fusion. In: 2024 8th CAA International Conference on Vehicular Control and Intelligence (CVCI). pp.~1--6 (2024). \doi{10.1109/CVCI63518.2024.10830035}

\bibitem{hao2025styledrive}
Hao, R., Jing, B., Yu, H., Nie, Z.: Styledrive: Towards driving-style aware benchmarking of end-to-end autonomous driving (2025), \url{https://arxiv.org/abs/2506.23982}

\bibitem{hasenjager2017personalization}
Hasenjäger, M., Wersing, H.: Personalization in advanced driver assistance systems and autonomous vehicles: A review. In: 2017 IEEE 20th International Conference on Intelligent Transportation Systems (ITSC). pp.~1--7 (2017). \doi{10.1109/ITSC.2017.8317803}

\bibitem{UniAD}
Hu, Y., Yang, J., Chen, L., Li, K., Sima, C., Zhu, X., Chai, S., Du, S., Lin, T., Wang, W., Lu, L., Jia, X., Liu, Q., Dai, J., Qiao, Y., Li, H.: Planning-oriented autonomous driving. In: 2023 IEEE/CVF Conference on Computer Vision and Pattern Recognition (CVPR). pp. 17853--17862 (2023). \doi{10.1109/CVPR52729.2023.01712}

\bibitem{jain2016brain4cars}
Jain, A., Koppula, H.S., Soh, S., Raghavan, B., Singh, A., Saxena, A.: Brain4cars: Car that knows before you do via sensory-fusion deep learning architecture (2016), \url{https://arxiv.org/abs/1601.00740}

\bibitem{DriveAdapter}
Jia, X., Gao, Y., Chen, L., Yan, J., Liu, P.L., Li, H.: Driveadapter: Breaking the coupling barrier of perception and planning in end-to-end autonomous driving. In: Proceedings of the IEEE/CVF International Conference on Computer Vision (ICCV). pp. 7953--7963 (October 2023)

\bibitem{ThinkTwice}
Jia, X., Wu, P., Chen, L., Xie, J., He, C., Yan, J., Li, H.: Think twice before driving: Towards scalable decoders for end-to-end autonomous driving. In: Proceedings of the IEEE/CVF Conference on Computer Vision and Pattern Recognition (CVPR). pp. 21983--21994 (June 2023)

\bibitem{Bench2Drive}
Jia, X., Yang, Z., Li, Q., Zhang, Z., Yan, J.: Bench2drive: Towards multi-ability benchmarking of closed-loop end-to-end autonomous driving. In: Advances in Neural Information Processing Systems. vol.~37, pp. 819--844. Curran Associates, Inc. (2024). \doi{10.52202/079017-0025}

\bibitem{DriveTransformer}
Jia, X., You, J., Zhang, Z., Yan, J.: Drivetransformer: Unified transformer for scalable end-to-end autonomous driving. In: International Conference on Learning Representations (ICLR) (2025)

\bibitem{VAD}
Jiang, B., Chen, S., Xu, Q., Liao, B., Chen, J., Zhou, H., Zhang, Q., Liu, W., Huang, C., Wang, X.: Vad: Vectorized scene representation for efficient autonomous driving. In: 2023 IEEE/CVF International Conference on Computer Vision (ICCV). pp. 8306--8316 (2023). \doi{10.1109/ICCV51070.2023.00766}

\bibitem{kim2024ndstneuraldrivingstyle}
Kim, D., Khalil, A., Nam, H., Kwon, J.: Ndst: Neural driving style transfer for human-like vision-based autonomous driving (2024), \url{https://arxiv.org/abs/2407.08073}

\bibitem{kou2025padriver}
Kou, G., Jia, F., Mao, W., Liu, Y., Zhao, Y., Zhang, Z., Yoshie, O., Wang, T., Li, Y., Zhang, X.: Padriver: Towards personalized autonomous driving (2025), \url{https://arxiv.org/abs/2505.05240}

\bibitem{li2025learningpersonalized}
Li, D., Li, C., Wang, Y., Ren, J., Wen, X., Li, P., Xu, L., Zhan, K., Jia, P., Lang, X., Xu, N., Zhao, H.: Learning personalized driving styles via reinforcement learning from human feedback (2025), \url{https://arxiv.org/abs/2503.10434}

\bibitem{metadrive}
Li, Q., Peng, Z., Feng, L., Zhang, Q., Xue, Z., Zhou, B.: { MetaDrive: Composing Diverse Driving Scenarios for Generalizable Reinforcement Learning}. IEEE Transactions on Pattern Analysis \& Machine Intelligence  \textbf{45}(03),  3461--3475 (Mar 2023). \doi{10.1109/TPAMI.2022.3190471}, \url{https://doi.ieeecomputersociety.org/10.1109/TPAMI.2022.3190471}

\bibitem{li2023personalized}
Li, S., Wei, C., Wu, G., Barth, M.J., Abdelraouf, A., Gupta, R., Han, K.: Personalized trajectory prediction for driving behavior modeling in ramp-merging scenarios. In: 2023 Seventh IEEE International Conference on Robotic Computing (IRC). pp.~1--4 (2023). \doi{10.1109/IRC59093.2023.00054}

\bibitem{li2022tradeoff}
Li, X.: Trade-off between safety, mobility and stability in automated vehicle following control: An analytical method. Transportation Research Part B: Methodological  \textbf{166},  1--18 (2022). \doi{https://doi.org/10.1016/j.trb.2022.09.003}, \url{https://www.sciencedirect.com/science/article/pii/S0191261522001461}

\bibitem{isEgoStatusAllYouNeed}
Li, Z., Yu, Z., Lan, S., Li, J., Kautz, J., Lu, T., Alvarez, J.M.: Is ego status all you need for open-loop end-to-end autonomous driving? In: 2024 IEEE/CVF Conference on Computer Vision and Pattern Recognition (CVPR). pp. 14864--14873 (2024). \doi{10.1109/CVPR52733.2024.01408}

\bibitem{liao2023driver}
Liao, X., Zhao, X., Wang, Z., Zhao, Z., Han, K., Gupta, R., Barth, M.J., Wu, G.: Driver digital twin for online prediction of personalized lane-change behavior. IEEE Internet of Things Journal  \textbf{10}(15),  13235--13246 (2023). \doi{10.1109/JIOT.2023.3262484}

\bibitem{liao2024review}
Liao, X., Zhao, Z., Barth, M.J., Abdelraouf, A., Gupta, R., Han, K., Ma, J., Wu, G.: A review of personalization in driving behavior: Dataset, modeling, and validation. IEEE Transactions on Intelligent Vehicles  \textbf{10}(2),  1241--1262 (2025). \doi{10.1109/TIV.2024.3425647}

\bibitem{ling2021towards}
Ling, J., Li, J., Tei, K., Honiden, S.: Towards personalized autonomous driving: An emotion preference style adaptation framework. In: 2021 IEEE International Conference on Agents (ICA). pp. 47--52 (2021). \doi{10.1109/ICA54137.2021.00015}

\bibitem{liu2025diverse}
Liu, W., Hu, W., Jing, W., Lei, L., Gao, L., Liu, Y.: Learning to model diverse driving behaviors in highly interactive autonomous driving scenarios with multiagent reinforcement learning. IEEE Systems Journal  \textbf{19}(1),  317--326 (2025). \doi{10.1109/JSYST.2025.3528976}

\bibitem{lu2019personalized}
Lu, C., Gong, J., Lv, C., Chen, X., Cao, D., Chen, Y.: A personalized behavior learning system for human-like longitudinal speed control of autonomous vehicles. Sensors  \textbf{19}(17) (2019). \doi{10.3390/s19173672}, \url{https://www.mdpi.com/1424-8220/19/17/3672}

\bibitem{natarajan2022adapt}
Natarajan, M., Akash, K., Misu, T.: Toward adaptive driving styles for automated driving with users' trust and preferences. In: 2022 17th ACM/IEEE International Conference on Human-Robot Interaction (HRI). pp. 940--944 (2022). \doi{10.1109/HRI53351.2022.9889313}

\bibitem{ilias2023intelligent}
Panagiotopoulos, I., Dimitrakopoulos, G.: Intelligent, in-vehicle autonomous decision-making functionality for driving style reconfigurations. Electronics  \textbf{12}(6) (2023). \doi{10.3390/electronics12061370}, \url{https://www.mdpi.com/2079-9292/12/6/1370}

\bibitem{pei2026safe}
Pei, S., Wang, Y., Zhu, Y., Sun, C., Li, Q., Zhao, Y., Tan, H.: Safe and stylized trajectory planning for autonomous driving via diffusion model (2026), \url{https://arxiv.org/abs/2602.04329}

\bibitem{transfuser}
Prakash, A., Chitta, K., Geiger, A.: Multi-modal fusion transformer for end-to-end autonomous driving. In: 2021 IEEE/CVF Conference on Computer Vision and Pattern Recognition (CVPR). pp. 7073--7083 (2021). \doi{10.1109/CVPR46437.2021.00700}

\bibitem{SimLingo}
Renz, K., Chen, L., Arani, E., Sinavski, O.: Simlingo: Vision-only closed-loop autonomous driving with language-action alignment. In: Proceedings of the Computer Vision and Pattern Recognition Conference (CVPR). pp. 11993--12003 (June 2025)

\bibitem{schrum2024maveric}
Schrum, M.L., Sumner, E., Gombolay, M.C., Best, A.: Maveric: A data-driven approach to personalized autonomous driving. IEEE Transactions on Robotics  \textbf{40},  1952--1965 (2024). \doi{10.1109/TRO.2024.3359543}

\bibitem{interfuser}
Shao, H., Wang, L., Chen, R., Li, H., Liu, Y.: Safety-enhanced autonomous driving using interpretable sensor fusion transformer. In: Proceedings of The 6th Conference on Robot Learning. Proceedings of Machine Learning Research, vol.~205, pp. 726--737. PMLR (14--18 Dec 2023), \url{https://proceedings.mlr.press/v205/shao23a.html}

\bibitem{sheng2022study}
Sheng, S., Pakdamanian, E., Han, K., Wang, Z., Feng, L.: A study on learning and simulating personalized car-following driving style. In: 2022 IEEE 25th International Conference on Intelligent Transportation Systems (ITSC). pp. 1208--1215 (2022). \doi{10.1109/ITSC55140.2022.9922548}

\bibitem{DriveLM}
Sima, C., Renz, K., Chitta, K., Chen, L., Zhang, H., Xie, C., Bei\ss{}wenger, J., Luo, P., Geiger, A., Li, H.: Drivelm: Driving with graph visual question answering. In: Computer Vision – ECCV 2024: 18th European Conference, Milan, Italy, September 29–October 4, 2024, Proceedings, Part LII. p. 256–274. Springer-Verlag, Berlin, Heidelberg (2024). \doi{10.1007/978-3-031-72943-0_15}, \url{https://doi.org/10.1007/978-3-031-72943-0_15}

\bibitem{speidel2019towards}
Speidel, O., Graf, M., Phan-Huu, T., Dietmayer, K.: Towards courteous behavior and trajectory planning for automated driving. In: 2019 IEEE Intelligent Transportation Systems Conference (ITSC). pp. 3142--3148 (2019). \doi{10.1109/ITSC.2019.8917033}

\bibitem{surmann2025multi}
Surmann, H., de~Heuvel, J., Bennewitz, M.: Multi-objective reinforcement learning for adaptable personalized autonomous driving (2025), \url{https://arxiv.org/abs/2505.05223}

\bibitem{tian2022personalized}
Tian, H., Wei, C., Jiang, C., Li, Z., Hu, J.: Personalized lane change planning and control by imitation learning from drivers. IEEE Transactions on Industrial Electronics  \textbf{70}(4),  3995--4006 (2023). \doi{10.1109/TIE.2022.3177788}

\bibitem{treiber2000idm}
Treiber, M., Hennecke, A., Helbing, D.: Congested traffic states in empirical observations and microscopic simulations. Phys. Rev. E  \textbf{62},  1805--1824 (Aug 2000). \doi{10.1103/PhysRevE.62.1805}, \url{https://link.aps.org/doi/10.1103/PhysRevE.62.1805}

\bibitem{wei2025pdb}
Wei, C., Qin, Z., Li, S., Zhang, Z., Zhao, X., Abdelraouf, A., Gupta, R., Han, K., Barth, M.J., Wu, G.: Pdb: Not all drivers are the same -- a personalized dataset for understanding driving behavior (2025), \url{https://arxiv.org/abs/2503.06477}

\bibitem{TCP}
Wu, P., Jia, X., Chen, L., Yan, J., Li, H., Qiao, Y.: Trajectory-guided control prediction for end-to-end autonomous driving: A simple yet strong baseline. In: Advances in Neural Information Processing Systems. vol.~35, pp. 6119--6132. Curran Associates, Inc. (2022)

\bibitem{yang2024drivingstyle}
Yang, R., Zhang, X., Fernandez-Laaksonen, A., Ding, X., Gong, J.: Driving style alignment for llm-powered driver agent. In: 2024 IEEE/RSJ International Conference on Intelligent Robots and Systems (IROS). pp. 11318--11324 (2024). \doi{10.1109/IROS58592.2024.10802629}

\bibitem{ADMLP}
Zhai, J.T., Feng, Z., Du, J., Mao, Y., Liu, J.J., Tan, Z., Zhang, Y., Ye, X., Wang, J.: Rethinking the open-loop evaluation of end-to-end autonomous driving in nuscenes (2023), \url{https://arxiv.org/abs/2305.10430}

\bibitem{zhang2023learn}
Zhang, Y., Xu, Q., Wang, J., Wu, K., Zheng, Z., Lu, K.: A learning-based discretionary lane-change decision-making model with driving style awareness. IEEE Transactions on Intelligent Transportation Systems  \textbf{24}(1),  68--78 (2023). \doi{10.1109/TITS.2022.3217673}

\bibitem{zhao2022personalized}
Zhao, Z., Wang, Z., Han, K., Gupta, R., Tiwari, P., Wu, G., Barth, M.J.: Personalized car following for autonomous driving with inverse reinforcement learning. In: 2022 International Conference on Robotics and Automation (ICRA). pp. 2891--2897 (2022). \doi{10.1109/ICRA46639.2022.9812446}

\bibitem{zhu2018personalized}
Zhu, B., Yan, S., Zhao, J., Deng, W.: Personalized lane-change assistance system with driver behavior identification. IEEE Transactions on Vehicular Technology  \textbf{67}(11),  10293--10306 (2018). \doi{10.1109/TVT.2018.2867541}

\bibitem{CARLALB2}
Zimmerlin, J., Beißwenger, J., Jaeger, B., Geiger, A., Chitta, K.: Hidden biases of end-to-end driving datasets (2024), \url{https://arxiv.org/abs/2412.09602}

\end{thebibliography}
